\documentclass[10pt,twocolumn,letterpaper]{article}

\usepackage[pagenumbers]{cvpr} 

\definecolor{cvprblue}{rgb}{0.21,0.49,0.74}
\usepackage[pagebackref,breaklinks,colorlinks,allcolors=cvprblue]{hyperref}
\usepackage{graphicx}
\usepackage{amsmath}
\usepackage{amssymb}
\usepackage{booktabs}
\usepackage{microtype}
\usepackage{colortbl}  
\usepackage{xcolor}
\usepackage{array}   
\usepackage{pifont}      
\usepackage{bbding}   
\usepackage{fontawesome}  
\usepackage{tcolorbox}
\usepackage{multirow}
\usepackage{wrapfig}
\usepackage{tcolorbox}

\newcommand{\name}{SpaceDrive}

\newcommand{\cmark}{\ding{51}}%
\def\paperID{*****} 
\def\confName{CVPR}
\def\confYear{2026}

\title{SpaceDrive: Infusing Spatial Awareness into VLM-based Autonomous Driving}

\author{
Peizheng~Li$^{*\,1,2}$, Zhenghao~Zhang$^{*\,1,4}$, David~Holtz$^{1}$, Hang~Yu$^{1,5}$, Yutong~Yang$^{1,6}$, \\
Yuzhi~Lai$^{2}$, Rui~Song$^{7}$, Andreas~Geiger$^{2,3}$, Andreas~Zell$^{2}$ \\
$^{1}$Mercedes-Benz AG, $^{2}$University of Tübingen, $^{3}$Tübingen AI Center,\\
$^{4}$TU Munich, $^{5}$Karlsruhe Institute of Technology, $^{6}$University of Stuttgart, $^{7}$UCLA\\
\tt \href{https://zhenghao2519.github.io/SpaceDrive_Page/}{https://zhenghao2519.github.io/SpaceDrive\_Page/}
}

\begin{document}

\twocolumn[{%
    \renewcommand\twocolumn[1][]{#1}%
    \maketitle
    \centering
    \includegraphics[width=\linewidth]{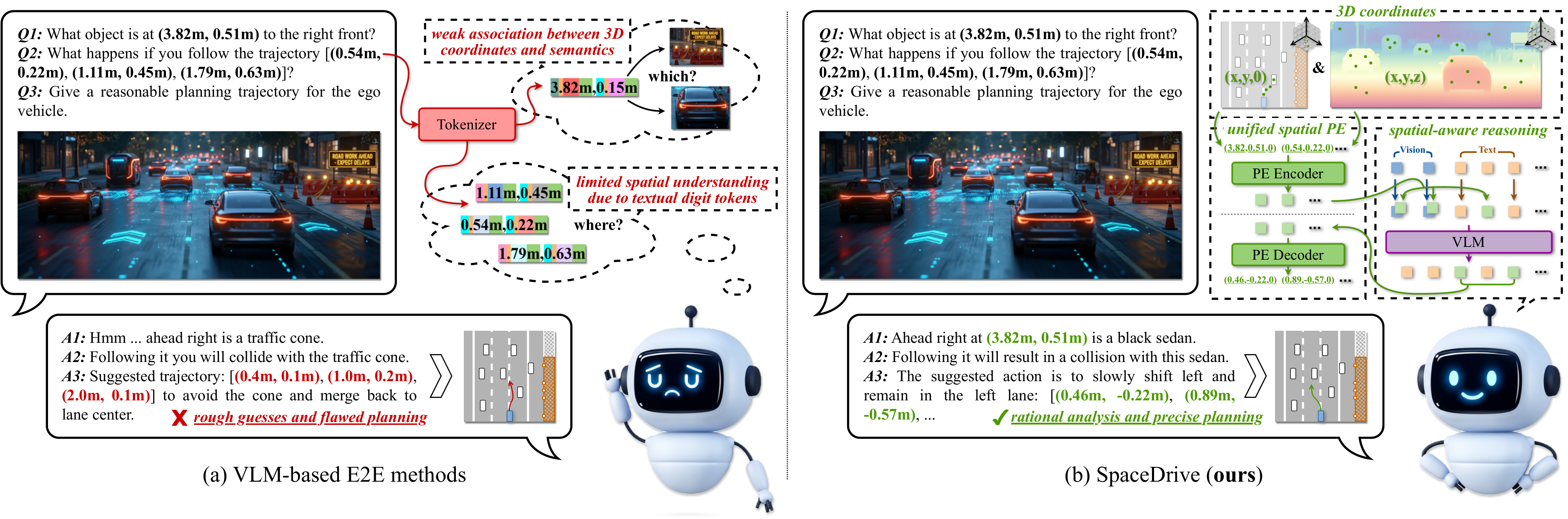}
    \vspace{-20pt}
    \captionof{figure}{
    \textbf{Spatial awareness in VLM-based end-to-end autonomous driving.}
    (a) Constrained by insufficient 3D pre-training and discrete token-wise encoding, existing end-to-end planners based on the VLM struggle to precisely ground, associate, and predict 3D spatial positions, limiting their planning capabilities.
    (b) Our proposed \name{} planner introduces a unified 3D coordinate encoding to replace the original VLM's textual digit tokens and augment visual features, achieving explicit association with 2D perspective semantics to enhance joint spatial reasoning for E2E planning.
    Compared to current VLM-based methods, it achieves state-of-the-art driving capability in the nuScenes open-loop evaluation and the second-best driving performance in the Bench2Drive closed-loop simulation.
    }
    \label{fig:teaser}
    \vspace{10pt}
}]

\begin{abstract}

End-to-end autonomous driving methods built on vision language models (VLMs) have undergone rapid development driven by their universal visual understanding and strong reasoning capabilities obtained from the large-scale pretraining. 
However, we find that current VLMs struggle to understand fine-grained 3D spatial relationships which is a fundamental requirement for systems interacting with the physical world.
To address this issue, we propose \name{}, a spatial-aware VLM-based driving framework that treats spatial information as explicit positional encodings (PEs) instead of textual digit tokens, enabling joint reasoning over semantic and spatial representations.
\name{} employs a universal positional encoder to all 3D coordinates derived from multi-view depth estimation, historical ego-states, and text prompts.
These 3D PEs are first superimposed to augment the corresponding 2D visual tokens.
Meanwhile, they serve as a task-agnostic coordinate representation, replacing the digit-wise numerical tokens as both inputs and outputs for the VLM.
This mechanism enables the model to better index specific visual semantics in spatial reasoning and directly regress trajectory coordinates rather than generating digit-by-digit, thereby enhancing planning accuracy.
Extensive experiments validate that \name{} achieves state-of-the-art open-loop performance on the nuScenes dataset and the second-best Driving Score of 78.02 on the Bench2Drive closed-loop benchmark over existing VLM-based methods.
Code is available at: https://github.com/zhenghao2519/SpaceDrive.

\begin{NoHyper}
    \def\thefootnote{*\,}\footnotetext{Equal contribution, names are sorted alphabetically. 
     Correspondence to: \{\href{mailto:peizheng.li@mercedes-benz.com}{peizheng.li}, \href{mailto:zhenghao.zhang@mercedes-benz.com}{zhenghao.zhang}\}@mercedes-benz.com.
     }

\end{NoHyper} 

\end{abstract}    

\section{Introduction}
\label{sec:intro}

Large-scale pre-trained VLMs are known for their vast knowledge bases and strong reasoning capabilities.
Leveraging VLMs to assist~\cite{tian2025drivevlm, jiang2024senna, pan2024vlp} or replace~\cite{sima2024drivelm, wang2025omnidrive, fu2025orion} traditional end-to-end (E2E) autonomous driving (AD) systems has therefore emerged as a prominent trend recently.
These systems typically reformulate AD functions into natural language, and flexibly perform scene understanding, motion prediction and trajectory planning based on semantic information extracted from images.
Compared to fixed modular designs~\cite{hu2023planning, jiang2023vad}, VLM-based E2E models promise to achieve superior generalization, addressing increasingly complex and dynamic driving scenarios.

However, current VLMs demonstrate clear limitations in 3D tasks such as geometric measurement and distance estimation~\cite{chen2024spatialvlm, yang2025thinking, wu2025spatial}, which are critical for autonomous driving.
This issue stems mainly from two primary factors, as illustrated in~\cref{fig:teaser}.a. 
First, the absence of 3D-data-based pre-training forces models to rely on inference from existing 2D knowledge.
When dealing with 3D coordinates, VLMs struggle to associate them with the corresponding objects and their 2D semantics, leading to ambiguous or even incorrect scene descriptions~\cite{zhang2025mpdrive}.
Second, language models inherently treat numerical processing as digit-by-digit classification.
This classification overlooks the inherent inter-digit proximity between numerical tokens and incorrectly averages the importance of different token positions~\cite{fei2025advancing}.

In autonomous driving, existing VLM-based planners either introduce task-specific embeddings tailored to individual downstream tasks~\cite{renz2025simlingo, fu2025orion} or represent waypoints as sequences of numeric tokens directly generated by the language model~\cite{sima2024drivelm, wang2025omnidrive}.
The former relies on specialized 3D fine-tuning, tying embeddings to particular tasks and domains and thus hindering a transferable, universal spatial representation that preserves VLM generalization.
The latter suffers from the aforementioned limitations in the numerical modeling ability of language models, results in inaccurate waypoint predictions.
However, an important but underemphasized aspect is that the Transformer architecture is \textbf{inherently capable of processing positional relationships between tokens}, which can be conceptualized as \textbf{spatial relationships between semantic features}~\cite{kerethinking}. 
Therefore, extending this capability to 3D spatial awareness becomes a natural and logical idea.

Inspired by this, we propose \textit{\name{}}, a spatial-aware VLM-based AD framework illustrated in~\cref{fig:teaser}.b, which incorporates a universal encoding for 3D positions to enhance spatial understanding and reasoning in VLMs.
Specifically, we first encode 3D coordinates derived from depth estimation and add them onto corresponding 2D visual tokens, establishing an explicit association between semantic features and 3D spatial locations.
Meanwhile, this 3D PE serves as a general coordinate representation, replacing either conventional coordinates in natural language or task-specific embeddings as the input and output of VLM.
Furthermore, for the output PE, we replace the original classification-based design with regression-based decoder and loss to address the numerical prediction deficiencies in language models.
Our framework also exhibits strong adaptability to various VLM base models and reasoning strategies, further underscoring its potential as a universal paradigm.

To directly validate the trajectory planning accuracy, we first conducted an open-loop evaluation.
Experiments on the nuScenes dataset~\cite{caesar2020nuscenes} demonstrate that \name{} achieves state-of-the-art performance among all VLM-based methods.
However, similarity-based open-loop planning evaluation is highly susceptible to dataset overfitting, offering only limited insight into the model's actual driving competence.
Therefore, we further validate our method on the closed-loop Bench2Drive~\cite{jia2024bench2drive} benchmark where we achieve a Driving Score of 78.02 (second-best in VLM-based planners), further confirming its capability to perform reasonable planning in dynamic and complex scenarios.

The contributions of this paper are as follows:
\begin{itemize}
    \item We identify fundamental limitations of current VLMs in 3D spatial reasoning and waypoint prediction, and propose \textit{\name{}}, a spatial-aware VLM-based AD framework with a universal 3D positional encoding that explicitly associates image semantics with 3D coordinates.
    \item \textit{\name{}} employs a shared 3D PE as a general coordinate representation to augment visual tokens and serve as the coordinate interface for language models, along with a regression-based decoder to enhance the end-to-end trajectory planning.
    \item Our framework achieves state-of-the-art performance in open-loop planning on nuScenes, while exhibits strong closed-loop planning capabilities under complex driving scenarios on the Bench2Drive benchmark.
\end{itemize}

\begin{figure*}[htbp]
    \centering
    \includegraphics[width=\linewidth]{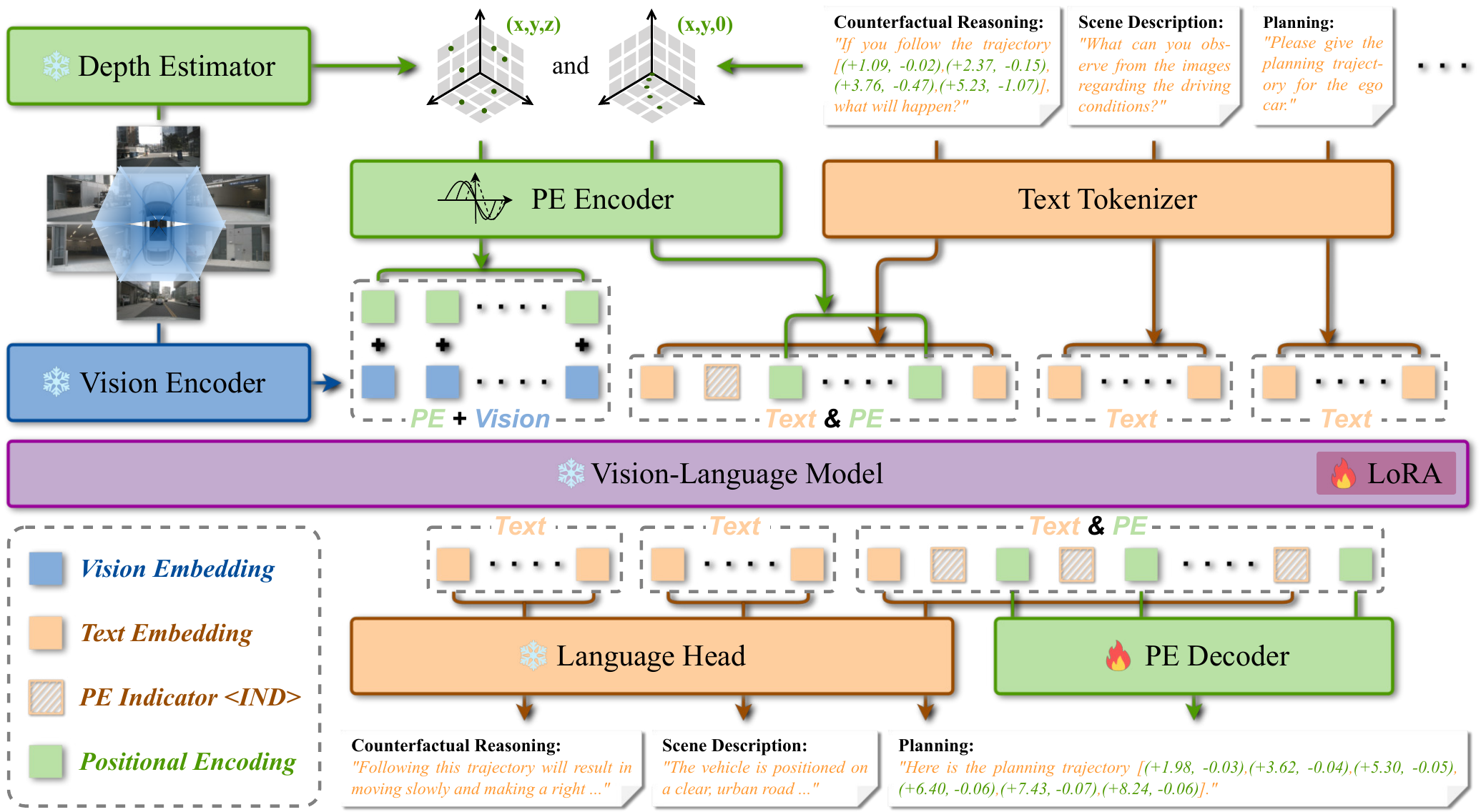}
    \vspace{-15pt}
    \caption{
    \textbf{\name{} framework.} 
    Beyond the base VLM, a frozen depth estimator predicts dense metric depths from surround-view images, which are projected into 3D coordinates and encoded by a universal PE encoder to augment visual tokens with spatial cues.
    BEV coordinates in text prompts are encoded by the same PE encoder, replacing the original coordinate tokens and preceded by the PE indicator $\langle\text{IND}\rangle$.
    At the output stage, the recognized PE is passed through a PE decoder to obtain the final coordinates for trajectory planning.
    }
    \label{fig:arch}
    \vspace{-15pt}
\end{figure*}
\section{Related Work}
\label{sec:related_work}

\noindent \textbf{End-to-End Autonomous Driving}
Over the past years, end-to-end autonomous driving has evolved from traditional modular stacks~\cite{wang2022detr3d, li2024bevformer, liu2022petr, li2025ago, zhang2024seflow, ijcai2023p120, ding2025tqd, yu2025hype, wu2026fusion, zhao2024balf} to fully differentiable, planning-oriented designs. 
After early methods like ST-P3~\cite{hu2022st} achieved joint optimization of perception and planning, UniAD~\cite{hu2023planning} unified the entire stack into a query-based framework, using planning supervision to regularize upstream tasks.
Building on this paradigm, follow-up studies~\cite{jiang2023vad, jia2023think, chen2024vadv2, zheng2024genad, weng2024drive, shen2025divide} achieved further improvements in planning efficiency and decision quality.
A key inflection came from AD-MLP~\cite{zhai2023rethinking} and BEV-Planner~\cite{li2024ego}, which exposed open-loop brittleness: simple ego-state priors can rival sophisticated stacks. 
This finding shifted attention toward closed-loop fidelity and benchmarks that align with driving quality, \eg Bench2Drive~\cite{jia2024bench2drive} and DriveE2E~\cite{yu2025drivee2e}, and stimulated numerous subsequent methods~\cite{jia2023think, zhangfuture, zimmerlin2024hidden, li2025end, li2025hydra, liao2025diffusiondrive, liu2025gaussianfusion, huangprioritizing, shang2025drivedpo} based on them.
Despite their strong performance, conventional E2E frameworks lack generalized scene understanding, thus struggling to handle complex and dynamic driving scenarios.

\noindent \textbf{Spatial Intelligence of VLMs}
Recently, spatial intelligence in VLMs has progressed from 2D relational heuristics to explicit 3D-aware reasoning~\cite{chen2024spatialvlm, fu2024scene, zhu2024llava, zheng2025video, wu2025spatial, lai2025fam, lai2025seer, wu2025stereoadapter, du2026unsupervised, zhao2026advances}.
This trend was initiated by SpatialVLM~\cite{chen2024spatialvlm}, which synthesized large-scale spatial Visual Question Answering (VQA) data to support both qualitative and quantitative spatial reasoning from 2D images.
Subsequent works injected 3D structure more directly into the modeling pipeline.
From integration of 3D features and positional embeddings in Scene-LLM~\cite{fu2024scene} and LLaVA-3D~\cite{zhu2024llava}, to dynamic and region-prompted spatial reasoning in Video-3D LLM~\cite{zheng2025video}, Spatial-MLLM~\cite{wu2025spatial}, and SR-3D~\cite{cheng20253d}, these works collectively advance language-guided 3D understanding, grounding, and planning.
Besides, dedicated benchmarks have standardized the evaluation.
VSI-Bench~\cite{yang2025thinking}  probes egocentric video-based visual–spatial intelligence with more than 5,000 QA pairs, while STI-Bench~\cite{li2025sti} stresses precise spatial–temporal estimation (pose, displacement, motion) across various scene setting.
These studies demonstrate the immense potential of VLMs in spatial-aware tasks and suggest clear benefits for the perception, prediction, and planning in autonomous driving.

\noindent \textbf{VLMs-Based Driving Agents}
Vision-language and multimodal LLMs have reshaped E2E driving by injecting priors, interactivity, and explicit reasoning into perception-prediction-planning. 
Early work such as DriveGPT4~\cite{xu2024drivegpt4} formulated driving as a language-conditioned sequence modeling, pairing video inputs with textual rationales to produce interpretable low-level controls. 
VLP~\cite{pan2024vlp} and DriveVLM~\cite{tian2025drivevlm} extended this direction by leveraging large vision-language models for scene understanding and trajectory generation, while DriveLM~\cite{sima2024drivelm} further strengthened structured reasoning via graph-structured VQA over driving scenes.
Recent methods~\cite{zhou2025autovla, zhou2025opendrivevla, yuan2024rag} achieve further enhancements in areas such as reinforcement learning, symbolic reasoning, and precise control. 
For example, OmniDrive~\cite{wang2025omnidrive} pursues holistic 3D grounding with counterfactual supervision, while ORION~\cite{fu2025orion} aligns reasoning and action spaces via a long-horizon QT-Former, an LLM reasoner, and a generative planner for strong closed-loop scores.
Concomitant with the methodological developments, corresponding benchmarks~\cite{shao2024lmdrive, sima2024drivelm, nie2024reason2drive, wang2025omnidrive, arai2025covla} have also arisen, primarily targeting on open-world reasoning and regulation compliance.
Nevertheless, existing VLM-based autonomous driving systems suffer from an inadequate treatment of 3D spatial awareness, a critical deficiency that forms the core focus of this paper.
\section{Method}
\label{sec:method}

As illustrated in~\cref{fig:arch}, we propose \name{}, a spatial-aware framework that enhances end-to-end planning through explicit injection of 3D information into the VLM architecture.
Specifically, the surrounding images are first encoded by a visual encoder, and then aligned to the language model's semantic space via a projector.
Meanwhile, these images are processed by a depth estimator to obtain absolute depths, which are converted into 3D positional encodings through a universal PE encoder. 
The visual tokens and their 3D PEs are then added element-wise, yielding spatially-aware visual tokens that serve as inputs to the VLM.
Besides, text prompts for various reasoning tasks are also fed into the VLM as text token inputs. 
Notably, Bird's-Eye-View (BEV) or 3D coordinates within these prompts are processed separately by the same PE encoder to generate universal PEs, replacing the corresponding original text tokens.
To avoid semantic confusion with other tokens, a predefined PE indicator is placed before each PE input and output.
During reasoning, these PEs leverage their intrinsic similarity for direct interaction and indexing of the spatially-aware visual tokens.
At the output stage, general textual outputs are decoded by the language head, while coordinate-related outputs are recognized and decoded by a dedicated PE decoder to produce accurate 3D coordinates for precise trajectory planning.

\subsection{Spatial Awareness in Perception}
\label{subsec:sa_in_perc}

A prerequisite for spatial intelligence is reliable 3D scene understanding, \ie establishing dense correlations between 2D perspective visual features and their 3D geometry. 

\noindent \textbf{Vision Encoding} 
A pretrained vision encoder $f_{vis.}$ first converts the $K$ multi-view images $\{I_k\}_{k=1}^{K}$ into $N$ patch tokens:
\begin{equation}
X_v = f_{vis.}(\{I_k\}) = \{x_p\}_{p=1}^{N}.
\end{equation}
Given that our primary goal is the explicit infusion of spatial awareness, the sparse and highly abstract features within Q-Former-style architectures~\cite{wang2025omnidrive} are fundamentally limited in directly associating with concrete 3D spatial locations. 
Furthermore, the efficacy of the Q-Former typically requires additional large-scale pre-training for vision-language alignment, largely reducing the adaptabilty of our framework.
Therefore, we keep using a simple MLP $g$ to densely align the visual and language feature spaces, consistent with general-purpose VLMs~\cite{liu2023visual,bai2025qwen2}:
\begin{equation}
H_v = g(X_v) = \{h_p\}_{p=1}^{N}.
\end{equation}

\noindent \textbf{Spatial Encoding}
To obtain 3D scene information, a pretrained depth estimator $f_{dep.}$ produces dense per-view absolute depth maps $D_k = f_{dep.}(I_k)$. 
To prioritize the foreground, for each patch $p$ with image-plane support $\mathcal{R}_p$ we assign the minimum depth $d_p = \min_{(u,v)\in \mathcal{R}_p} D_k(u,v)$ as its corresponding depth.
With the per-camera calibration matrix $\mathcal{P}_k$, we project the patch center $(u_p,v_p)$ to 3D as $\mathbf{c}_p = \mathcal{P}_k^{\dagger}[u_p, v_p, d_p, 1]z{\top}$ to obtain explicit metric coordinates.
Each $\mathbf{c}_p = (x_p^{3D}, y_p^{3D}, z_p^{3D})$ is then encoded into a universal 3D positional encoding via a PE encoder.
To minimize confusion with the existing RoPE~\cite{su2024roformer} used in the VLM, we opt for a 3D sine-cosine positional encoding extending the standard 1D formulation dimension-wise:
\begin{equation}
\begin{aligned}
    \phi(\mathbf{c}_p)
    &=\big[\phi_x(x_p^{3D}),\phi_y(y_p^{3D}),\phi_z(z_p^{3D})\big]\in\mathbb{R}^{dim}, \text{with} \\
    \phi_a(p_a)
    &=\begin{cases}
    \sin(\tfrac{p_a}{20000^{2i/d_a}}),\\[1pt]
    \cos(\tfrac{p_a}{20000^{2i/d_a}}),
    \end{cases}
    i=0,\dots,\lfloor\tfrac{d_a}{2}\rfloor-1,\\[-1pt]
    d_x&=d_y=\lceil\tfrac{{dim}}{3}\rceil,
    d_z={dim}-d_x-d_y.
    \label{eq:3D_PE}
\end{aligned}
\end{equation}
for spatial dimension $a\in\{x,y,z\}$ and total PE width ${dim}$.

\noindent \textbf{Spatial Token Injection} 
Prior works~\cite{wang2025omnidrive, fu2025orion} inject learnable 3D cues within or before the vision-language projector, yielding only implicit geometry.
In contrast, we explicitly add metric 3D coordinates information $\phi(\mathbf{c}_p)$ on top of modality-aligned visual tokens $h_p$ after the MLP $g$.
This design enables later reuse of the same PE $\phi(\cdot)$ for coordinates from text prompts, allowing the model to directly index spatially grounded visual features and strengthening downstream spatial reasoning, as further discussed in~\cref{subsec:sa_in_reas}.

It is worth noting that direct additive injection of $\phi(\mathbf{c}_p)$ shifts the token norm distribution away from the pretrained VLM regime. 
To mitigate this, we introduce a learnable normalization factor $\alpha_{PE}$ shared across all 3D PEs, simply
\begin{equation}
\tilde{H_v} = \{\tilde{h_p}\}_{p=1}^{N}, \, \tilde{h}_p = h_p + \alpha_{PE}\, \phi(\mathbf{c}_p).
\end{equation}

\subsection{Spatial Awareness in Reasoning}
\label{subsec:sa_in_reas}

Existing VLMs exhibit strong general 2D multimodal reasoning yet remain deficient in explicit 3D spatial inference:
\begin{enumerate}
    \item Insufficient pretraining on metric 3D data and spatial reasoning tasks confines current VLMs mainly to abstract 2D reasoning~\cite{zhu2024llava}, yielding poor estimation of inter-object spatial relations, physical extent, and distances. 
    \item The classification-based numerical prediction in existing language models often prioritizes fitting data distributions while neglecting the inherent affinity between numerical symbols and their sequential order~\cite{fei2025advancing}, thereby degrading precision in continuous waypoint predictions.
\end{enumerate}
Alternatively, existing methods introduce task-specific queries and decode explicit 3D coordinates from them using MLPs~\cite{renz2025simlingo}, generative modules~\cite{fu2025orion} or attention layers~\cite{zhu2024llava}.
Although partially mitigating the above limitations, the resulting tokens lack unified spatial semantics and thus transfer poorly across tasks.
In contrast, we reuse the previously defined 3D PE $\phi(\mathbf{c})$ as a universal spatial representation. 
This choice enforces representational consistency between perception and reasoning, improving accuracy of coordinate handling and estimation within the VLM.

\noindent \textbf{Encoding of Coordinates in Text Prompts} 
During tokenization of input text prompts, we scan the text sequence $\{t_i\}_{i=1}^{L}$ for substrings $\mathcal{S}$ expressing spatial coordinates. 
For each detected coordinate expression we extract its numeric values as a vector $\mathbf{c}_r = (x_r,y_r,z_r)$.
The same 3D positional encoder $\phi(\cdot)$ as in~\cref{subsec:sa_in_perc} is then applied to obtain a corresponding spatial token $\phi(\mathbf{c}_r) \in \mathbb{R}^{dim}$, which replaces the original sequence of numeric tokens corresponding to that coordinate.
Each input PE is preceded by a specifically defined token, $\langle\text{IND}\rangle$, serving as the PE identifier (for simplicity, $\langle\text{IND}\rangle$ will be omitted in subsequent descriptions and formulations).
The adjusted text token inputs are as follows:
\begin{equation}
\tilde{H_t}=\{\tilde{h}_i\}_{i=1}^L, \, \tilde{h}_i=\begin{cases}\phi(\mathbf{c}_r) & i\in\mathcal{S}_r \\ \mathrm{Tokenizer}(t_i) & \text{otherwise}\end{cases} .
\end{equation}
A special case arises for BEV coordinates (\eg trajectory waypoints), where we set all $z$-axis components in the PE $\phi(\mathbf{c}_r)$ to 0 so that they do not contribute to subsequent attention calculations.

\noindent \textbf{Encoding of the Ego Status} 
It has been verified that ego state inputs are highly effective for trajectory planning~\cite{zhai2023rethinking, li2024ego}.
Existing approaches typically encode all state variables (\eg pose, velocity, acceleration) simply into a single vector embedding $\mathbf{e}_{\text{ego}}\in\mathbb{R}^{dim}$, mostly also augmented with BEV features to obscure explicit metric structure. 
Thanks to our unified spatial representation, we instead encode the historical ego waypoints via the same $\phi(\cdot)$ employed before, \ie $\{\phi(\mathbf{c}^{ego}_\tau)\}_{\tau=-T}^{1}$.
It will then be fed into the language model together with $\mathbf{e}_{\text{ego}}$ as explicit spatial-temporal conditioning for accuracy trajectory planning.

\noindent \textbf{Decoding of Text with Coordinates}
At the output stage, the VLM produces a sequence of embeddings $\{\mathbf{e}_j\}_{j=1}^{J}$. 
A standard language head $W_{\text{lang}}$ maps each $\mathbf{e}_j$ to a distribution over the textual vocabulary $\mathcal{V}$ for ordinary decoding. 
Additionally, we utilize the previously defined $\langle\text{IND}\rangle$ to signal a forthcoming coordinate emission, extending the original $W_{\text{lang}}$ to $W'_{\text{lang}}$, \ie
\begin{equation}
y_j = \arg\max_{y \in \mathcal{V}'} \big(W'_{\text{lang}} \mathbf{e}_j\big)_y,\, 
\mathcal{V}'=\mathcal{V}\cup\{\langle\text{IND}\rangle\}.
\end{equation}
If $y_j \neq \langle\text{IND}\rangle$ the text token is emitted normally. 
When $y_j=\langle\text{IND}\rangle$, $\mathbf{e}_j$ remains in the language context and the subsequent output state $\mathbf{e}_{j+1}$ is routed to a PE decoder $\psi(\cdot)$ to produce metric coordinates:
\begin{equation}
\hat{\mathbf{c}} = \psi(\mathbf{e}_{j+1}), \, \hat{\mathbf{c}} \in \mathbb{R}^3
\end{equation}
This mechanism yields precise BEV trajectory waypoints (omitting the $z$-coordinate) while preserving autoregressive continuity for surrounding text. 
Because the composite sinusoidal encoding $\phi(\mathbf{c})$ is not analytically invertible (phase and frequency aliasing across dimensions), $\psi(\cdot)$ is set as a fully learnable MLP and trained to regress ground-truth coordinates. 
The shared use of $\phi(\cdot)$ in both perception and reasoning ensures that $\psi(\cdot)$ operates over embeddings already aligned with unified spatial PEs, improving coordinate fidelity and trajectory planning accuracy.

\subsection{Loss Function}
\label{subsec:loss}
A typical training objective combines language modeling $\mathcal{L}_{\text{LM}}$ (applied to all text outputs) with coordinate regression $\mathcal{L}_{\text{reg.}}$ (applied to all coordinate outputs, such as waypoints):
\begin{equation}
\mathcal{L} = \mathcal{L}_{\text{LM}} + \mathcal{L}_{\text{reg.}}(\hat{\mathbf{c}}, \mathbf{c}),
\end{equation}
where $\mathcal{L}_{\text{reg.}}$ may vary with the type of the decoder $\psi(\cdot)$ and the adopted trajectory generation strategy.
For the basic MLP decoder, we adopt the Huber loss for coordinate regression.

\section{Experiments}
\label{sec:experiments}

\subsection{Experimental Settings}
\label{subsec:exp_setting}

\begin{table*}[t]
    \caption{
    \textbf{Open-loop planning results on nuScenes~\cite{caesar2020nuscenes}.} 
    \name{}+ denotes the adoption of the ego planner input.
    Methods categorized as \textcolor{gray}{Hybrid Paradigm} simultaneously stack traditional and VLM-based approaches, and are thus incomparable.
    Results are highlighted in \textbf{bold} and \underline{underline} for the best and second-best performance among VLM-based methods, respectively.
    All results in this table follow the evaluation protocol of OmniDrive~\cite{wang2025omnidrive} and ORION~\cite{fu2025orion}. 
    More methods based on other metrics are provided in the supplementary materials.
    }
    \vspace{-5pt}
    \label{tab:open_loop_benchmark}
    \centering
    \resizebox{0.8\linewidth}{!}{
    \begin{tabular}{l|cc|cccc|cccc|cccc}
    \toprule
    \multirow{2}{*}{Method} &
    \multicolumn{2}{c|}{Ego Status} &
    \multicolumn{4}{c|}{L2 (m) $\downarrow$} & 
    \multicolumn{4}{c|}{Collision (\%) $\downarrow$} &
    \multicolumn{4}{c}{Intersection (\%) $\downarrow$} \\
    &BEV &Planner& 1s & 2s & 3s &\cellcolor{gray!30}Avg. & 1s & 2s & 3s& 
    \cellcolor{gray!30}Avg. & 1s & 2s & 3s &\cellcolor{gray!30}Avg.\\
    
    \midrule
    \multicolumn{15}{l}{\textit{Traditional Modular Paradigm}} \\
    \midrule
    
    ST-P3~\cite{hu2022st} & - & - & 1.33 & 2.11 & 2.90 & \cellcolor{gray!30}2.11 & 0.23 & 0.62 & 1.27 & \cellcolor{gray!30}0.71 & 2.53 & 8.17 & 14.40 & \cellcolor{gray!30}8.37 \\
    UniAD~\cite{hu2023planning} &\cmark & \cmark & 0.20 & 0.42 & 0.75 & \cellcolor{gray!30}0.46 & 0.02 & 0.25 & 0.84 & \cellcolor{gray!30}0.37 & 0.20 & 1.33 & 3.24 & \cellcolor{gray!30}1.59  \\
    VAD-Base~\cite{jiang2023vad} &\cmark& \cmark& 0.17 & 0.34 & 0.60 & \cellcolor{gray!30}0.37 & 0.04 & 0.27 & 0.67 & \cellcolor{gray!30}0.33 & 0.21 & 2.13 & 5.06 & \cellcolor{gray!30}2.47  \\
    AD-MLP~\cite{zhai2023rethinking} & - & \cmark& 0.15 & 0.32 & 0.59 & \cellcolor{gray!30}0.35 & 0.00 & 0.27 & 0.85 & \cellcolor{gray!30}0.37 & 0.27 & 2.52 & 6.60 & \cellcolor{gray!30}2.93\\
    BEV-Planner++~\cite{li2024ego} &\cmark &\cmark & 0.16 & 0.32 & 0.57 & \cellcolor{gray!30}0.35 & 0.00 & 0.29 & 0.73 & \cellcolor{gray!30}0.34 & 0.35 & 2.62 & 6.51 & \cellcolor{gray!30}3.16 \\
    UAD~\cite{guo2025end} &\cmark &\cmark & 0.13 & 0.28 & 0.48 & \cellcolor{gray!30}0.30 & 0.00 & 0.19 & 0.61 & \cellcolor{gray!30}0.27 & 0.13 & 1.08 & 2.89 & \cellcolor{gray!30}1.37 \\
    Drive-WM~\cite{wang2024driving} &\cmark &\cmark & 0.43 & 0.77 & 1.20 & \cellcolor{gray!30}0.80 & 0.10 & 0.21 & 0.48 & \cellcolor{gray!30}0.26 & - & - & - & \cellcolor{gray!30}- \\
    
    \midrule
    \multicolumn{15}{l}{\textit{VLM-based Paradigm}} \\
    \midrule
    
    EMMA~\cite{hwang2024emma} & - & - & \textbf{0.14} & \textbf{0.29} & \underline{0.54} & \cellcolor{gray!30}\textbf{0.32} & - & - & - & \cellcolor{gray!30}- & - & - & - & \cellcolor{gray!30}- \\
    RDA-Driver~\cite{huang2024making} &\cmark &\cmark & 0.23 & 0.73 & 1.54 & \cellcolor{gray!30}0.80 & \textbf{0.00} & \textbf{0.13} & 0.83 & \cellcolor{gray!30}0.32 & - & - & - & \cellcolor{gray!30}-  \\
    DriveVLM~\cite{tian2025drivevlm} & - & \cmark & 0.18 & 0.34 & 0.68 & \cellcolor{gray!30}0.40 & 0.10 & 0.22 & \textbf{0.45} & \cellcolor{gray!30}\underline{0.27} & - & - & - & \cellcolor{gray!30}- \\
    ORION~\cite{fu2025orion}  &\cmark & - & 0.17 & \underline{0.31} & 0.55 & \cellcolor{gray!30}0.34 & 0.05 & 0.25 & 0.80 & \cellcolor{gray!30}0.37 & - & - & - & \cellcolor{gray!30}-  \\
    OmniDrive-Q~\cite{wang2025omnidrive} & - & - & 1.15 & 1.96 & 2.84 & \cellcolor{gray!30}1.98 & 0.80 & 3.12 & 7.46 & \cellcolor{gray!30}3.79 & 1.66 & 3.86 & 8.26 & \cellcolor{gray!30}4.59\\
    OmniDrive-Q++~\cite{wang2025omnidrive} &\cmark &\cmark & \textbf{0.14} & \textbf{0.29} & 0.55 & \cellcolor{gray!30}\underline{0.33} & \textbf{0.00} & \textbf{0.13} & 0.78 & \cellcolor{gray!30}0.30 & \underline{0.56} & \underline{2.48} & \underline{5.96} & \cellcolor{gray!30}\underline{3.00} \\

    \midrule
    
    \name{}~(\textbf{ours}) & - & - & 1.06 & 1.79 & 2.55 & \cellcolor{gray!30}1.80 & 0.35 & 1.33 & 3.97 & \cellcolor{gray!30}1.88 & 0.96 & 3.38 &  8.28 & \cellcolor{gray!30}4.21 \\
    \name{}+~(\textbf{ours}) & - & \cmark & \underline{0.15} & \textbf{0.29} & \textbf{0.51} & \cellcolor{gray!30}\textbf{0.32} & \underline{0.04} & \underline{0.18} & \underline{0.49} & \cellcolor{gray!30}\textbf{0.23} & \textbf{0.22} & \textbf{0.80} & \textbf{2.79} & \cellcolor{gray!30}\textbf{1.27} \\

    \midrule
    \multicolumn{15}{l}{\textcolor{gray}{\textit{Hybrid Paradigm}}} \\
    \midrule

    \textcolor{gray}{VLP~\cite{pan2024vlp}} & \textcolor{gray}{\cmark} & \textcolor{gray}{-} & \textcolor{gray}{0.30} & \textcolor{gray}{0.53} & \textcolor{gray}{0.84} & \cellcolor{gray!30}\textcolor{gray}{0.55} & \textcolor{gray}{0.01} & \textcolor{gray}{0.07} & \textcolor{gray}{0.38} & \cellcolor{gray!30}\textcolor{gray}{0.15} & \textcolor{gray}{-} & \textcolor{gray}{-} & \textcolor{gray}{-} & \cellcolor{gray!30}\textcolor{gray}{-} \\
    \textcolor{gray}{ReAL-AD~\cite{lu2025real}} & \textcolor{gray}{\cmark} & \textcolor{gray}{-} & \textcolor{gray}{0.30} & \textcolor{gray}{0.48} & \textcolor{gray}{0.67} & \cellcolor{gray!30}\textcolor{gray}{0.48} & \textcolor{gray}{0.07} & \textcolor{gray}{0.10} & \textcolor{gray}{0.28} & \cellcolor{gray!30}\textcolor{gray}{0.15} & \textcolor{gray}{-} & \textcolor{gray}{-} & \textcolor{gray}{-} & \cellcolor{gray!30}\textcolor{gray}{-} \\
    \textcolor{gray}{DriveVLM-Dual~\cite{tian2025drivevlm}} & \textcolor{gray}{\cmark} & \textcolor{gray}{-} & \textcolor{gray}{0.15} & \textcolor{gray}{0.29} & \textcolor{gray}{0.48} & \cellcolor{gray!30}\textcolor{gray}{0.31} & \textcolor{gray}{0.05} & \textcolor{gray}{0.08} & \textcolor{gray}{0.17} & \cellcolor{gray!30}\textcolor{gray}{0.10} & \textcolor{gray}{-} & \textcolor{gray}{-} & \textcolor{gray}{-} & \cellcolor{gray!30}\textcolor{gray}{-} \\
    \textcolor{gray}{SOLVE-VLM~\cite{chen2025solve}} & \textcolor{gray}{\cmark} & \textcolor{gray}{-} & \textcolor{gray}{0.13} & \textcolor{gray}{0.25} & \textcolor{gray}{0.47} & \cellcolor{gray!30}\textcolor{gray}{0.28} & \textcolor{gray}{0.00} & \textcolor{gray}{0.16} & \textcolor{gray}{0.43} & \cellcolor{gray!30}\textcolor{gray}{0.20} & \textcolor{gray}{-} & \textcolor{gray}{-} & \textcolor{gray}{-} & \cellcolor{gray!30}\textcolor{gray}{-} \\
    \textcolor{gray}{Senna~\cite{jiang2024senna}} & \textcolor{gray}{\cmark} & \textcolor{gray}{-} & \textcolor{gray}{0.11} & \textcolor{gray}{0.21} & \textcolor{gray}{0.35} & \cellcolor{gray!30}\textcolor{gray}{0.22} & \textcolor{gray}{0.04} & \textcolor{gray}{0.08} & \textcolor{gray}{0.13} & \cellcolor{gray!30}\textcolor{gray}{0.08} & \textcolor{gray}{-} & \textcolor{gray}{-} & \textcolor{gray}{-} & \cellcolor{gray!30}\textcolor{gray}{-} \\
    \bottomrule
    \end{tabular}}
    \vspace{-15pt}
\end{table*}

\begin{table}[t]
    \caption{
    \textbf{Closed-loop planning results on Bench2Drive~\cite{jia2024bench2drive}.}
    Results are highlighted in \textbf{bold} and \underline{underline} for the best and second-best performance among VLM-based methods, respectively.
    }
    \label{tab:closed_loop_benchmark}
    \vspace{-5pt}
    \centering
    \resizebox{0.8\linewidth}{!}{
    \begin{tabular}{l|cc}
    \toprule
    \multirow{2}{*}{Method} &
    \multicolumn{2}{c}{Closed-loop Metric} \\
    & Driving Score $\uparrow$ & Success Rate(\%) $\uparrow$ \\
    
    \midrule
    \multicolumn{3}{l}{\textit{Traditional Modular Paradigm}} \\
    \midrule
    
    AD-MLP~\cite{zhai2023rethinking} & 18.05 & 0.00 \\ 
    UniAD-Base~\cite{hu2023planning} & 45.81 & 16.36 \\ 
    VAD-Base~\cite{jiang2023vad} & 42.35 & 15.00 \\ 
    MomAD~\cite{song2025don} & 44.54 & 16.71 \\ 
    GenAD~\cite{zheng2024genad} & 44.81 & 15.90 \\ 
    SparseDrive~\cite{sun2025sparsedrive} & 47.38 & 17.72 \\ 
    UAD~\cite{guo2025end} & 49.22 & 20.45 \\ 
    WoTE~\cite{li2025end} & 61.71 & 31.36 \\ 
    ThinkTwice~\cite{jia2023think} & 62.44 & 37.17 \\ 
    DriveTransformer-L~\cite{jia2025drivetransformer} & 63.46 & 38.60 \\ 
    DriveAdapter~\cite{jia2023driveadapter} & 64.22 & 42.08 \\ 
    HiP-AD~\cite{tang2025hipad} & 86.77 & 69.09 \\ 
    AlignDrive~\cite{wu2026aligndrive} & 89.07 & 73.18 \\
     
    \midrule
    \multicolumn{3}{l}{\textit{VLM-based Paradigm}} \\
    \midrule
    
    ReAL-AD~\cite{lu2025real} & 41.17 & 11.36 \\ 
    Dual-AEB~\cite{zhang2025dual} & 45.23 & 10.00 \\ 
    VDRive~\cite{guo2025vdrive} & 66.25 & 50.51 \\ 
    StuckSolver~\cite{bao2025large} & 70.89 & 50.01 \\ 
    DriveMoE~\cite{yang2025drivemoe} & 74.22 & 48.64 \\ 
    ETA~\cite{hamdan2025eta} & 74.33 & 48.33 \\ 
    VLR-Drive~\cite{kong2025vlr} & 75.01 & 50.00 \\ 
    ORION~\cite{fu2025orion} & 77.74 & 54.62 \\ 
    SimLingo~\cite{renz2025simlingo} & \textbf{85.07} & \textbf{67.27} \\ 

    \midrule

    \name{}+~(\textbf{ours}) & \underline{78.02} & \underline{55.11} \\ 
    
    \bottomrule
    \end{tabular}}
    \vspace{-15pt}
\end{table}

\noindent \textbf{Dataset and Metrics}
The nuScenes dataset~\cite{caesar2020nuscenes} comprises 1{,}000 urban driving scenes (train/val/test: $700/150/150$) with full-stack \(360^\circ\) sensing (6 cameras, 1 LiDAR, 5 radars). 
For open-loop planning we predict 6 waypoints within a 3~s horizon and evaluate (i) waypoint displacement (L2) error, (ii) Collision rate (fraction of future timestamps overlapping with any dynamic agent), and (iii) Intersection rate (fraction of timestamps intruding into non-drivable map regions).

Bench2Drive~\cite{jia2024bench2drive} is a closed-loop planning benchmark emphasizing interactive scenarios (merging, overtaking, yielding, emergency negotiation) in a deterministic
CARLA V2~\cite{dosovitskiy2017carla} simulator.
Our closed-loop evaluation adopts the official protocol of 220 short routes, covering 44 interactive scenarios, with 5 distinct routes defined for each scenario.
Closed-loop metrics include Driving Score (route progress penalized by safety infractions) and Success Rate (percentage of scenarios completed without terminal violation).
All reported results adopt identical horizon, temporal sampling, footprint inflation, and map definitions for fair comparison.

\noindent \textbf{Implementation Details}
Our model adopts Qwen2.5-VL-7B~\cite{bai2025qwen2} as the base VLM.
We finetune the core LLM using LoRA~\cite{hu2022lora} with the rank of 16, while keeping the original vision encoder and vision-language projector frozen.
Unidepthv2-ViT-L~\cite{piccinelli2025unidepthv2} is chosen as our default depth estimation module without additional finetuning.
For the open-loop evaluation on nuScenes, the model is trained for 6 epochs on 8\(\times\)A100 80GB GPUs with a batch size of 8.
The learning rate is set to 1e-4 and cosine annealing is used to ensure stable training.
The input resolution is resized to $640\times640$.
For the closed-loop evaluation, the model is trained for 12 epochs using the same training setup as the open-loop evaluation.
For VQA training and evaluation, we adopt the data and settings utilized in OmniDrive~\cite{wang2025omnidrive}.
Further details are provided in the supplementary materials.

\subsection{Quantitative Results}
\label{subsec:quantitative_results}
\noindent \textbf{Open-loop Planning}
To directly validate the impact of spatial awareness on VLM’s coordinate regression, we first conducted the open-loop planning evaluation on the nuScenes dataset.
As shown in~\cref{tab:open_loop_benchmark}, \name{}+ achieves the SOTA performance across all reported metrics, consistently surpassing existing VLM-based methods. 
The lowest L2 error (0.32) indicates that coordinate-level regression allows closer adherence to expert driving trajectories. 
Simultaneously, the markedly reduced Collision (0.23\%) and Intersection (1.27\%) rates further show that \name{}+ excels not only at fitting ground truth but also enhances autonomous driving safety comprehensively through superior spatial understanding and reasoning.

Notably, our method does not include the BEV features widely adopted in existing pipelines. 
This provides evidence that a unified positional encoding is sufficient for 3D spatial modeling within VLM-oriented autonomous driving, obviating dense BEV representations.
Considering the sensitivity of open-loop metrics to the integration of ego status~\cite{li2024ego}, we further report the variant without ego status inputs, \ie \name{}. 
In this setting, our method also surpasses its codebase (OmniDrive~\cite{wang2025omnidrive}) across all dimensions (L2: -0.18, Collision: -1.91\%, Intersection: -0.38\%), validating the effectiveness of explicitly injecting 3D spatial information.

\noindent \textbf{Closed-loop Planning}
In~\cref{tab:closed_loop_benchmark} we conduct closed-loop evaluation to establish a comprehensive and reliable assessment of planning performance. 
While our codebase (OmniDrive) attains competitive open-loop metrics, its text-only planning paradigm fails drastically in closed-loop simulation (Under 10\% Success Rate). 
Empirically, its predicted trajectories collapse into near-linear paths with unstable heading oscillations. 
This substantiates our hypothesis that pure natural-language trajectory generation primarily fits data priors rather than learning a controllable driving pattern.
More comparisons are provided in the supplementary materials.

By introducing explicit spatial tokens, $\text{\name{}}+$ achieves 78.02 Driving Score and 55.11\% Success Rate, ranking as the second-best VLM-based method (notably, SimLingo~\cite{renz2025simlingo} employs extensive data augmentation via Action Dreaming).
These gains indicate that injecting structured 3D positional information is sufficient to unlock strong closed-loop planning within a VLM-oriented framework. 

\subsection{Qualitative Results}
\label{subsec:qualitative_results}
\begin{figure*}[t]
    \centering
    \includegraphics[width=\linewidth]{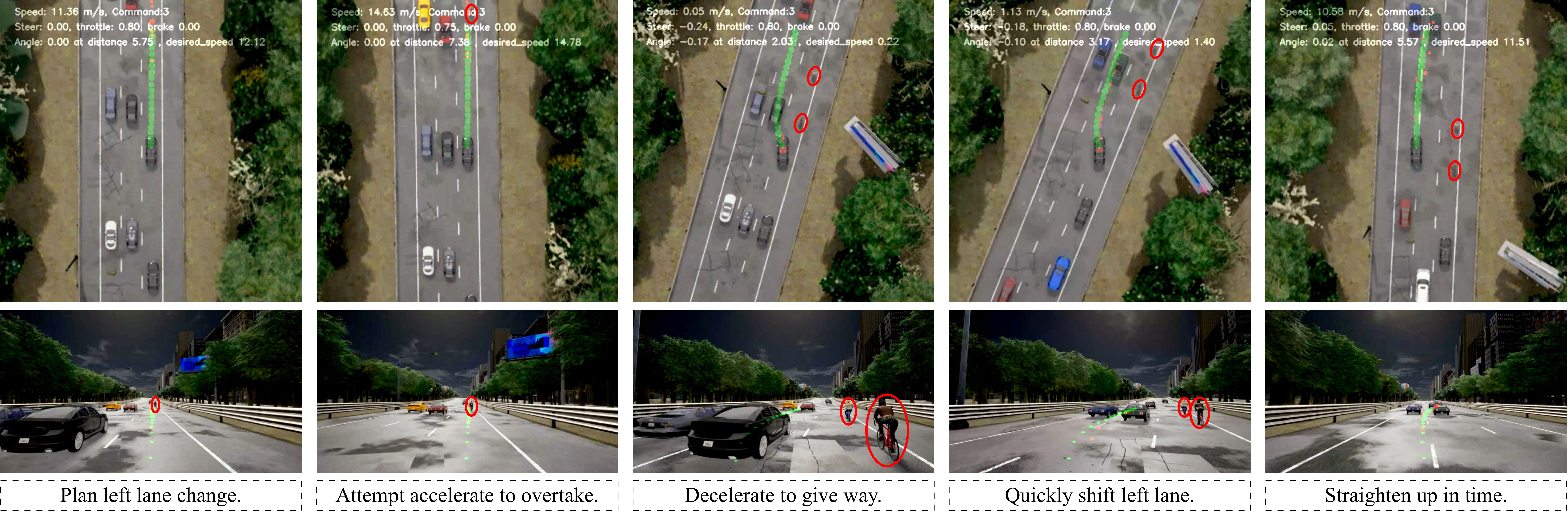}
    \vspace{-15pt}
    \caption{
    \textbf{Qualitative results of closed-loop evaluation on Bench2Drive~\cite{jia2024bench2drive}.} 
    \textcolor{green}{Green} and \textcolor{pink}{pink} dots represent path and speed waypoints, respectively. 
    \textcolor{red}{Red circles} indicate cyclists ahead that the vehicle needs to avoid.
    Parameters such as speed and steering wheel angle can be found in the figures.
    }
    \label{fig:planning_vis}
    \vspace{-10pt}
\end{figure*}

\Cref{fig:planning_vis} illustrates a representative Bench2Drive scenario in which the ego vehicle is required to avoid collision with two cyclists ahead. 
The planner first accelerates to attempt overtaking and lane changing.
After observing that the adjacent vehicle does not yield, our spatial-aware \name{}+ detects sufficient rearward clearance in the target lane and opts to decelerate to create a safe insertion gap. 
Once the opening emerges, it executes a decisive lateral maneuver. 
As the lane change nears completion, the model infers from its ego state and surrounding vehicle positions that rapid heading re‑alignment is necessary to avoid drifting out of lane boundaries. 
This case exemplifies that the injected 3D spatial encoding enables \name{}+ to adapt its strategy to evolving scene geometry and generate safety‑aware plans.

\subsection{Ablation Studies}
\label{subsec:ablation_study}
Our approach focuses on endowing models with spatial awareness rather than tailoring them for open- or closed-loop planning. 
Therefore, considering that closed-loop performance may be influenced by training strategies, PID controller tuning, and other pipeline heuristics, we conducted our ablations under open-loop settings (excluding the ego status to prevent overfitting) to ensure a fair comparisons.

\begin{table}[t]
    \caption{
    \textbf{Ablation of positional encoding.} 
    Here, $\phi(\mathbf{c}_r)$, $\phi(\mathbf{c}_p)$ and $\phi(\mathbf{c}^{ego}_t)$ denote the usage of spatial positional encodings for textual coordinate inputs, 3D coordinates corresponding to vision tokens and past ego locations, respectively.
    }
    \label{tab:abla_PE_embed}
    \vspace{-5pt}
    \centering
    \resizebox{\linewidth}{!}{
    \begin{tabular}{c|ccc|c|c|c} 
    \toprule
    Exp. & $\phi(\mathbf{c}_p)$ & $\phi(\mathbf{c}_r)$ & $\phi(\mathbf{c}^{ego}_\tau)$ & Avg. L2 $\downarrow$ & Avg. Col. $\downarrow$ & Avg. Int. $\downarrow$  \\ 
    \midrule
    \multicolumn{7}{l}{\textit{\name{}}} \\
    \midrule
    1 & & &  &2.51 & 4.53 & 6.77 \\
    2 & \cmark & & & 1.88  & 2.45 & 2.36 \\
    3 & & \cmark & & 2.42 & 5.06 & 8.94 \\
    4 & \cmark & \cmark & &1.80  &1.88  & 4.21 \\
    \midrule
    \multicolumn{7}{l}{\textit{\name{}+}} \\
    \midrule
    5 &  & & & 0.41   & 0.60 & 4.40 \\
    6 & \cmark & \cmark  & & 0.33 & 0.23 & 1.32 \\
    7 & \cmark & \cmark & \cmark & 0.32 & 0.23 & 1.27\\
    \bottomrule
    \end{tabular}}
    \vspace{-15pt}
\end{table}

\noindent \textbf{Positional Encoding} 
We compare the effect of injecting explicit positional encodings into different modules of the VLM-based planner in~\cref{tab:abla_PE_embed}. 
First, adding spatial encoding to vision tokens (Exp.~2 vs.~1) yields substantial improvements on all metrics, \ie -0.63 L2, -2.08\% Collision and -4.14\% Intersection.
This is largely attributable to the enhanced spatial understanding achieved by supplying 3D geometric context alongside 2D image features.
Meanwhile, the gains from replacing the textual coordinate with PE, as demonstrated in Exp.~3, are relatively smaller, likely because the PE lacks the bridge to associate with 2D semantic space in pretrained VLMs. 
However, when a unified positional encoding is applied to both vision and textual coordinate streams (Exp.~4 vs.~1; Exp.~6 vs.~5), planning performance improves regardless of the use of ego status, underscoring the value of a shared spatial representation. 
Finally, with ego status enabled, injecting past ego positions using the same $\phi(\cdot)$ (Exp.~7 vs.~6) further reduces L2 error and Intersection rates. 
This indicaties that the benefits coming from consistent spatial tokens are stable and reliable, facilitating spatial understanding and reasoning in VLMs.

\begin{table}[t]
    \caption{
    \textbf{Ablation of PE encoder \& decoder.
    } 
    \textcolor{gray}{Gray} indicates that only 4929 out of 5119 output samples are semantically reasonable.
    }
    \label{tab:abla_PE_en_decoder}
    \vspace{-5pt}
    \centering
    \resizebox{\linewidth}{!}{
    \begin{tabular}{c|cc|c|c|c} 
    \toprule
    Exp. & Encoder $\phi(\cdot)$ & Decoder $\psi(\cdot)$ & Avg. L2 $\downarrow$ & Avg. Col. $\downarrow$ & Avg. Int. $\downarrow$ \\ 
    \midrule
    4 & Sine-Cosine & Coordinate-wise & 1.80  &1.88  & 4.21 \\
    \midrule
    8 & MLP & Coordinate-wise & 1.96 & 3.17 & 6.76 \\
    \textcolor{gray}{9} & \textcolor{gray}{RoPE} & \textcolor{gray}{Coordinate-wise} & \textcolor{gray}{1.93} & \textcolor{gray}{3.71} & \textcolor{gray}{11.40} \\
    \midrule
    10 & Sine-Cosine & Sine-Cosine & 1.87 & 2.62 & 9.20  \\
    11 & Sine-Cosine & Task-specific & 1.93 & 2.41 & 5.58 \\
    \bottomrule
    \end{tabular}}
    \vspace{-10pt}
\end{table}

\noindent \textbf{PE Encoder \& Decoder}
\Cref{tab:abla_PE_en_decoder} compares different encoders and decoders of PE. 
Using a Sine–Cosine encoder yields a translation‑invariant encodability.
It assists the attention layers in recovering inter‑token spatial relations, giving clear gains over a fully learnable MLP encoder (Exp.~4 vs.~8). 
Although RoPE shares the same property as the additive Sine–Cosine encoding, it leads to a large performance degradation and instability due to the confusion with existing RoPE used in the base VLM (Exp.~4 vs.~9). 
On the decoder side, numerically inverting Sine–Cosine encodings to precise coordinates is ill‑posed (only coarse interpolation is possible), and the output embedding space of a large VLM is typically not fully aligned with its input space. 
These factors make a learnable coordinate‑wise MLP decoder preferable, as reflected in its lower L2 error (1.80 in Exp.~4 vs 1.87 in Exp.~10).
A common paradigm in VLM planners is to use a task‑specific embedding and decode an entire trajectory from it.
This limits reuse across tasks and forces retraining when objectives change. 
The comparison between experiments 4 and 11 shows that jointly decoding multiple waypoints from a single embedding underperforms the coordinate‑wise strategy in all metrics, which predicts each waypoint conditioned on shared spatial tokens. 

\begin{table}[t]
    \caption{
    \textbf{Ablation of PE normalization.}
    \textcolor{gray}{Gray} indicates that only 2421 out of 5119 output samples are semantically reasonable.
    }
    \label{tab:abla_PE_norm}
    \vspace{-5pt}
    \centering
    \resizebox{\linewidth}{!}{
    \begin{tabular}{c|c|c|c|c} 
    \toprule
    Init. $\alpha_{PE}$ & Learnable & Avg. L2 $\downarrow$ & Avg. Collision $\downarrow$ & Avg. Intersection $\downarrow$ \\ 
    \midrule
    1 &  & 2.34  & 3.63 & 8.46\\
    0.1 &  & 2.43  & 3.79 & 9.42 \\
    \textcolor{gray}{0.02} &  & \textcolor{gray}{2.22} & \textcolor{gray}{2.71} & \textcolor{gray}{10.17} \\
    \midrule
    1 & \cmark   &1.82 & 2.04& 4.62\\
    0.1 & \cmark   & 1.80 & 1.88 & 4.21 \\
    0.02 &\cmark  &1.86 & 2.03 & 5.42\\
    \bottomrule
    \end{tabular}}
    \vspace{-15pt}
\end{table}

\noindent \textbf{PE Normalization}
In transformer-based VLMs, the embedding norm directly modulates its relative importance in the attention operations. 
Consequently, the norm of the PE can largely affect training stability and planning accuracy. 
In~\cref{tab:abla_PE_norm} we compare performance under different fixed initialization scales $\alpha_{PE}$. 
For Qwen-VL, smaller static $\alpha_{PE}$ values lead to consistent degradation across all open-loop metrics and even semantic instability, implying that excessively small PE norms result in negligible attention scores and hinder convergence.
Since the optimal PE scale differs between foundation models and may shift over training, we promote $\alpha_{PE}$ to a learnable parameter. 
This adaptive normalization, while mitigating semantic instability, produces a about -0.5\,m reduction in Avg. L2 together with marked moderation in Collision and Intersection rates.
This outcome validates the importance of a learnable normalization coefficient for the unified 3D PE.

More ablations and experiments regarding VQA, depth estimator, etc., are provided in the supplementary materials.

\begin{table}[t]
    \caption{
    \textbf{\name{} based on different VLM foundations.}
    }
    \label{tab:base_vlm}
    \vspace{-5pt}
    \centering
    \resizebox{\linewidth}{!}{
    \begin{tabular}{c|c|c|c|c} 
    \toprule
    Method & Base VLM & Avg. L2 $\downarrow$ & Avg. Collision $\downarrow$ & Avg. Intersection $\downarrow$ \\ 
    \midrule
    \multirow{2}{*}{\name{}} & LLaVA & 1.82 & 2.44 & 4.08 \\
    & Qwen-VL & 1.80 & 1.88 & 4.21 \\
    \midrule
    \multirow{2}{*}{\name{}+} & LLaVA & 0.31 & 0.23 & 1.42 \\
    & Qwen-VL & 0.32 & 0.23 & 1.27 \\
    \bottomrule
    \end{tabular}}
    \vspace{-15pt}
\end{table}

\subsection{Adaptability}
\label{subsec:adaptability}
Our proposed 3D spatial representation also demonstrates excellent adaptability.
First, it attains comparable performance on both Qwen-VL and LLaVA (See~\cref{tab:base_vlm}), indicating that the strong performance arise from unified spatial reasoning rather than backbone-specific biases. 
Injecting spatial awareness also preserves compatibility with inference‑time reasoning enhancements. 
For example, without ego state inputs we observe distributional degeneration of predicted trajectories, \ie mode collapse. Augmenting \name{} with lightweight chain‑of‑thought prompting (CoT) stabilizes waypoint diversity and reduces collapse without retraining.
Furthermore, we can also adapt the PE decoder into a VAE-based generative model, which has also been proven effective in improving closed‑loop robustness~\cite{fu2025orion}.
More comparisons are provided in the supplementary materials.
All these results collectively show that our spatial encoding is a general booster for spatial reasoning and E2E planning across VLM foundations and inference paradigms.
\section{Conclusion}
\label{sec:conclusion}
In this paper, we presented \name{}, a spatial-aware VLM-based end-to-end autonomous driving framework, that unifies 3D spatial awareness and multimodal reasoning through a universal positional encoding interface. 
The approach infuses metric 3D positional encodings into visual tokens, textual coordinate mentions, and historical ego states, and decodes coordinates with a regression head instead of digit-wise language generation. 
In this way, the model achieves superior trajectory planning performance compared to methods relying solely on natural language fitting.
Besides, this design reduces reliance on task-specific embeddings, unleashing the generalization capabilities of pre-trained VLMs for space-relevant reasoning.
Extensive experiments show state-of-the-art open-loop planning results on nuScenes across displacement, collision, and map compliance metrics, as well as strong closed-loop performance on Bench2Drive, substantially narrowing the gap between VLM planners and specialized end-to-end baselines. 
Ablations confirm the benefit of unified spatial tokens, coordinate-wise decoding, and learnable normalization of positional encodings. 
Meanwhile, \name{} also achieves good transferability across VLM backbones and remains adaptable to language reasoning strategies.
Limitations include the absence of explicit uncertainty handling, and no exploitation of multi-frame temporal memory mechanisms, which also represent potential directions for future research.
Nevertheless, we believe the proposed unified spatial representation offers a principled path toward more reliable, generalizable spatial-aware VLM-driven autonomy.

\section*{Acknowledgement}
This work is a result of the joint research project STADT:up (Förderkennzeichen 19A22006O). 
The project is supported by the German Federal Ministry for Economic Affairs and Climate Action (BMWK), based on a decision of the German Bundestag. 
The author is solely responsible for the content of this publication.

{
    \small
    \bibliographystyle{ieeenat_fullname}
    \bibliography{main}
}

\clearpage
\setcounter{page}{1}
\maketitlesupplementary

\appendix
\setcounter{section}{0}
\setcounter{table}{0}
\setcounter{figure}{0}
\setcounter{equation}{0}

\newcommand{\appendixlabel}[1]{\Alph{#1}}

\renewcommand{\thesection}{\appendixlabel{section}}
\renewcommand{\thetable}{\appendixlabel{table}}
\renewcommand{\thefigure}{\appendixlabel{figure}}
\renewcommand{\theequation}{\appendixlabel{equation}}

The main content of this supplementary material is organized as follows: 
\begin{itemize}
\item \Cref{sec:addl_imple_details}: More implementation details of our method; 
\item \Cref{sec:addl_exp_ana}: Additional experiments and ablation studies;
\item \Cref{sec:addl_vis}: Additional visualization comparisons and qualitative analysis.
\end{itemize}

\section{Additional Implementation Details}
\label{sec:addl_imple_details}

\subsection{\name{} Framework}
\label{subsec:framework_details}
To ensure seamless adaptability, our method avoids any model-specific customization and fully preserves the original image preprocessing, patchification strategy, text tokenization, and chat template used by each base VLM model.
Given the shape of the preprocessed visual patches, the depth map is resized accordingly using min-pooling (\eg patch shapes of $6\times 24\times 24$ for LLaVA-1.5-7B~\cite{liu2023visual} and $6\times 23\times 23$ for Qwen2.5-VL-7B~\cite{bai2025qwen2} in our configuration).
For the coordinate parsing of the text inputs, We employ a strict regex rule matching the format ``(\textit{X}, \textit{Y})'' or ``(\textit{X}, \textit{Y}, \textit{Z})''.
Substrings matching this pattern (\eg ``(7.5, -3.2)'') are extracted and extended as the 3D coordinates (\eg $(7.5, -3.2, 0)$) and converted to 3D PE.
Any numeric string failing this match (\eg ``3'') is treated as a standard textual token.
This strict separation prevents ambiguity between spatial coordinates and general numerical values, ensuring that only spatial data triggers the PE en/decoding.
Newly introduced tokens, such as $\langle\text{IND}\rangle$, are set as learnable and appended to the frozen input embedding layer and output language-model head.
The PE decoder for coordinates is implemented as a standard two-layer MLP with the same hidden dimensionality as the base VLM. 
In all experiments, we set the seed to 888. The LoRA configurations are listed in~\cref{tab:lora_config}.

\subsection{Training Details}
\label{subsec:training_details}

\noindent \textbf{Open-loop Planning}
For open-loop planning, we follow prior works and use 6 future trajectory points sampled at 2~Hz over a 3-second horizon as ground truth supervision.
As emphasized in previous studies~\cite{zhai2023rethinking, li2024ego}, strong open-loop planning performance can be achieved using only ego-status.
To rigorously validate the effectiveness of our framework, the standard \name{} variant intentionally excludes motion dynamics and high-level driving commands (\eg ``go straight'', ``turn right'') from its inputs.
In this configuration, the model performs trajectory planning exclusively from image observations, enabling a clean evaluation of the spatial reasoning capability brought by our design. 
The variant \name{}+ includes the current commands and ego dynamics of past 2 frames that are widely used in other works~\cite{wang2025omnidrive,fu2025orion}.

For VQA training and evaluation, we adopt the dataset provided by OmniDrive~\cite{wang2025omnidrive}, which includes scene description, attention, counterfactual reasoning, planning, as well as other general conversations. 
Consistent with the implementation of OmniDrive, the other VQA tasks are appended subsequent to the trajectory planning task to ensure semantic stability.

\noindent \textbf{Closed-loop Planning}
Inspired by SimLingo~\cite{renz2025simlingo}, we augment the supervision of 6 trajectory points with 20 additional path waypoints, uniformly spaced at 1-meter intervals.
In this setup, the trajectory points serve two purposes: estimating the target speed and identifying the appropriate waypoint for the target direction.
This leads to generally more stable steering regardless of whether the ego vehicle is moving or not.
Two PID contollers are applied to determine acceleration and steering, respectively.
During training, we use a subset of SimLingo routes containing 3600 episodes with PDM-lite as the expert driver.

\begin{table}
    \caption{
    \textbf{LoRA configurations for VLM fine-tuning.}
    \label{tab:lora_config}
    }
    \centering
    \resizebox{\linewidth}{!}{
    \begin{tabular}{l|cccc}
    \toprule
    Setting & Rank ($r$) & Alpha ($\alpha$) & Dropout & Target Modules\\
    \midrule
    Value & 16 & 16 & 0.05 & q\_proj, k\_proj, v\_proj, o\_proj  \\
    \bottomrule
    \end{tabular}
    }
\end{table}

\subsection{Learnable Parameter Convergence}
\label{subsec:PE_convergence}
We report that the learnable norm factor $\alpha_{PE}$ (initialized at 0.1) generally converges to the range of $\left[ 0.085, 0.11 \right]$ across different backbones and settings. 
For instance, in the default open-loop QwenVL-based setting, it stabilizes at 0.087.

\section{Additional Experiments and Analyses}
\label{sec:addl_exp_ana}

\begin{table}
    \caption{
    \textbf{Counterfactual reasoning comparison in the open-loop planning (without ego status).}
    P and R here stand for Precision and Recall.
    Results are highlighted in \textbf{bold} and \underline{underline} for the best and the second-best performance.
    \label{tab:supp_abla_VQA}
    }
    \centering
    \resizebox{\linewidth}{!}{\begin{tabular}{l|cc|cc|cc|cc}
    \toprule
    \multirow{2}{*}{Method} & \multicolumn{2}{c|}{Safe} & \multicolumn{2}{c|}{Red Light} & \multicolumn{2}{c|}{Collision} & \multicolumn{2}{c}{Drivable Area} \\
    & P & R & P & R & P & R & P & R \\
    \midrule
    BEV-MLP & 70.2 & 17.3 & 48.7 & 53.6 & 31.1 & 70.4 & 32.4 & 56.6 \\
    Omni-L~\cite{wang2025omnidrive} & \textbf{72.1} & \underline{58.0} & \underline{59.2} & \underline{63.3} & \underline{34.3} & \underline{71.3} & \underline{49.1} & \textbf{59.2} \\
    Omni-Q~\cite{wang2025omnidrive} & \underline{70.7} & 49.0 & 57.6 & 58.3 & 32.3 & \textbf{72.6} & 48.5 & \underline{58.6} \\
    \midrule
    SpaceDrive~(\textbf{ours}) & 65.7 & \textbf{63.6} & \textbf{70.3} & \textbf{72.7} & \textbf{37.5} & 66.4 & \textbf{55.0} & 37.0 \\
    \bottomrule
    \end{tabular}
    }
\end{table}

\subsection{VQA for Counterfactual Reasoning}
\label{subsec:vqa}

As aforementioned, to validate the spatial reasoning capabilities of \name{}, we conduct counterfactual reasoning experiments following the setting in OmniDrive~\cite{wang2025omnidrive}, as presented in~\cref{tab:supp_abla_VQA}.
In this evaluation, keywords such as ``safety'', ``collision'', ``running a red light'', and ``out of the drivable area'' are extracted from the VQA outputs and compared against ground truth keywords to compute Precision and Recall. 
The results demonstrate that our framework achieves superior performance across the majority of metrics, \eg a Recall of 63.6\% in the safety task.
It is particularly noteworthy that, without any specific prompt engineering for the dialogue, the mere incorporation of the unified 3D spatial representation enables significantly higher Precision in tasks demanding rigorous spatial understanding, such as Collision (37.5\%) and Drivable Area (55.0\%). 
This further confirms our \name{} possesses strong spatial reasoning capabilities.

\subsection{More Ablation Studies}
\label{subsec:addl_ablation_study}

\begin{table}
    \caption{
    \textbf{Ablation of depth estimator.}
    \label{tab:supp_abla_depth_est}
    }
    \centering
    \resizebox{\linewidth}{!}{
    \begin{tabular}{c|c|c|c} 
    \toprule
    $f_{dep.}$ & Avg. L2 $\downarrow$ & Avg. Collision $\downarrow$ & Avg. Intersection $\downarrow$ \\ 
    \midrule
    DepthAnythingV2~\cite{yang2024depth}  &1.76 & 1.95 & 3.96\\
    UniDepthV2~\cite{piccinelli2025unidepthv2}  &1.80  &1.88  & 4.21\\
    \bottomrule
    \end{tabular}}
\end{table}

\noindent \textbf{Depth Estimator}
In~\cref{tab:supp_abla_depth_est}, we compare the influence of different pre-trained depth estimator on the planning performance. 
DepthAnythingV2~\cite{yang2024depth} and UniDepthV2~\cite{piccinelli2025unidepthv2} are selected as representative examples of relative and metric depth estimation models, respectively. 
We observe that both variants perform similarly on the L2 error metric and Collision rate, which are the most reliable indicator of planning performance. 
This suggests that the effectiveness of our \name{} is independent of a specific pre-trained depth model, implicitly demonstrating the adaptability of our framework. 
Notably, LiDAR-based depth ground truth (GT) is inherently sparse and lacks valid depth values in regions such as the sky, necessitating manual definition. 
Together with factors like camera distortion and projection error, GT-based comparisons are unreliable and thus excluded from the comparison.

\begin{table}
    \caption{
    \textbf{Ablation of pooling strategy.}
    \label{tab:supp_abla_pooling}
    }
    \centering
    \resizebox{\linewidth}{!}{
    \begin{tabular}{c|c|c|c}
    \toprule
    Pooling Strategy & Avg. L2 $\downarrow$ & Avg. Collision $\downarrow$ & Avg. Intersection $\downarrow$\\
    \midrule
    Min & 1.80 & 1.88 & 4.21    \\
    Average & 1.83 & 1.88 & 4.08    \\
    Median & 1.83 & 2.17 & 4.64    \\
    \bottomrule
    \end{tabular}}
\end{table}

\noindent \textbf{Depth Pooling Strategy} 
Min pooling in our default setting may be sensitive to outliers, yet the compact patch resolution renders such artifacts empirically rare. 
In fact, every pooling strategy has its inherent limitations: 
average pooling blurs object boundaries, while median pooling induces intra-object depth discontinuities of adjacent patches.
In contrast, min-depth is relatively reliable and safety-critical as it preserves the closest obstacles. 
\cref{tab:supp_abla_pooling} validates that min pooling yields the lowest L2 error and collision rate.

\begin{table}
    \caption{
    \textbf{Ablation of LoRA rank.} 
    Learn. Par. is the abbreviation for the number of LoRA parameters when selecting Qwen2.5-VL-7B as the base VLM.
    \label{tab:supp_abla_lora_rank}
    }
    \centering
    \resizebox{\linewidth}{!}{
    \begin{tabular}{c|c|c|c|c} 
    \toprule
    Rank & Learn. Par. & Avg. L2 $\downarrow$ & Avg. Collision $\downarrow$ & Avg. Intersection $\downarrow$ \\ 
    \midrule
    16 & 10.09M &1.80  &1.88  & 4.21 \\
    64 &40.37M &1.88 & 2.13 & 4.08 \\
    128 & 80.74M & 1.82& 2.25 & 4.68 \\
    \bottomrule
    \end{tabular}}
\end{table}

\noindent \textbf{LoRA Rank}
\Cref{tab:supp_abla_lora_rank} presents a comparison of different LoRA~\cite{hu2022lora} ranks in the VLM during fine-tuning. Benefiting from our universal spatial positional encoding, the coordinate regression process in the language model is simplified. 
Utilizing only low-rank fine-tuning (rank 16) achieves the optimal overall result (L2 error of 1.80, Collision rate of 1.88\%, and Intersection rate of 4.21\%). 
While increasing the rank to 128 substantially raises the number of learnable parameters from 10.09M  to 80.74M, it fails to improve the planning accuracy and, instead, leads to a degradation in Collision and Intersection rates. 
We attribute this to the excessive degrees of training freedom in the high-rank adapter, which hinders the convergence. 
The above comparison further demonstrates that our method not only offers stronger planning reliability but also maintains parameter efficiency.

\begin{table}
    \caption{
    \textbf{Ablation of PE frequency.}
    \label{tab:supp_abla_PE_freq}
    }
    \centering
    \resizebox{\linewidth}{!}{
    \begin{tabular}{c|c|c|c} 
    \toprule
    Frequency & Avg. L2 $\downarrow$ & Avg. Collision $\downarrow$ & Avg. Intersection $\downarrow$ \\ 
    \midrule
    $1000^{-2i/d_a}$ & 1.78 & 2.01 & 3.93 \\
    $10000^{-2i/d_a}$ & 1.83 & 1.83& 3.18 \\
    $20000^{-2i/d_a}$ &1.80  &1.88  & 4.21 \\
    \bottomrule
    \end{tabular}}
\end{table}

\noindent \textbf{PE Frequency}
\Cref{tab:supp_abla_PE_freq} investigates the base frequency of the Sin-Cos PE, which impacts both encoding resolution and smoothness. 
Utilizing a smaller frequency base (corresponding to a higher frequency) introduces a larger phase shift between adjacent positions but leads to positional aliasing at long distances, thereby inhibiting the representation of far-field positions. 
As shown in the comparison, setting the base to 1000 enhances local resolution and achieves the lowest L2 error of 1.78.
However, distant coordinates exhibit near-random phase characteristics, which compromises overall safety (leading to worse Collision and Intersection rates). 
Conversely, an excessively large base (\eg 20000) generates smoother, more stable encodings over long distances but diminishes local discriminative capability. 
Compared to the original base of 10000, the resulting L2 error reduction is less pronounced, at only $-0.03$, but the collision rate increase is negligible. 
Overall, comparing all variants reveals that the influence of different PE frequencies is relatively limited and non-decisive. 
We finally adopt 20000 as our PE frequency base.

\begin{table}
    \caption{
    \textbf{Ablation of regression loss.}
    \label{tab:supp_abla_reg_loss}
    }
    \centering
    \resizebox{\linewidth}{!}{
    \begin{tabular}{c|c|c|c} 
    \toprule
    $\mathcal{L}_{\text{reg.}}$ & Avg. L2 $\downarrow$ & Avg. Collision $\downarrow$ & Avg. Intersection $\downarrow$ \\ 
    \midrule
    MAE  & 1.86 & 1.82 & 5.73 \\
    MSE  & 1.82 & 2.14 & 6.22\\
    Huber Loss & 1.80  & 1.88  & 4.21 \\
    \bottomrule
    \end{tabular}}
\end{table}

\noindent \textbf{Regression Loss}
In~\cref{tab:supp_abla_reg_loss}, we compares different regression losses for trajectory prediction.
MAE provides robustness to outliers but yields the worst L2 and intersection metrics, suggesting insufficient pressure on medium-scale errors. 
MSE reduces L2 compared to MAE, but its quadratic growth on large residuals makes optimization more sensitive to outliers, leading to noticeably higher collision and intersection rates. 
Huber loss strikes a balance between them and achieves the best L2 error together with markedly improved safety metrics.
So we adopt Huber loss as our final regression objective.

\subsection{Robustness to Depth Estimation Errors} 
\label{subsec:robustness}

\begin{table}
    \caption{
    \textbf{Robustness to Depth Estimation Errors.}
    \label{tab:robustness}
    }
    \centering
    \resizebox{\linewidth}{!}{
    \begin{tabular}{cc|c|c|c} 
    \toprule
    \multicolumn{2}{c|}{Depth Noise} & Avg. L2 $\downarrow$ & Avg. Collision $\downarrow$ & Avg. Intersection $\downarrow$ \\ 
    \midrule
    \multirow{3}{*}{Training \& Inference} & w. -5\% Global Shift   & 1.86      & 1.80        & 5.22 \\
     & w. 5\% Global Shift & 1.86 & 1.93 & 4.41 \\
     & w. 5\% Random Noise & 1.83 & 2.16 & 4.44 \\
    \midrule
    \multirow{3}{*}{Inference only} & w. -2.5\% Global Shift & 1.80 & 1.89 & 4.22 \\
     & w. 2.5\% Global Shift & 1.80 & 1.86 & 4.24 \\
     & w. 2.5\% Random Noise & 1.80 & 1.89 & 4.17 \\ 
    \bottomrule
    \end{tabular}}
\end{table}

We present experiments when injecting relative depth noise ($\pm$~5\% Global Shift or 5\% Random Noise) in inference or training \& inference.
As shown in~\cref{tab:robustness}, injecting noise \textit{during inference} results in negligible performance drop (\eg Avg. L2 remains 1.80).
This confirms that \name{} is robust to depth inaccuracies. 
It relies on \textit{3D spatial structure} rather than precise metric depth, allowing VLM's semantic understanding to compensate for noise.
Conversely, noise injection \textit{during training} slightly degrades performance.
This verifies that depth information is actively utilized in training, but the learned policy generalizes well against test-time perturbations.
In summary, \name{} treats depth as a geometric guide for attention, not strong geometric priors.

\subsection{Comprehensive Benchmark}
\label{subsec:full_benchmarks}

\begin{table*}[t]
    \caption{
    \textbf{Open-loop planning results on nuScenes~\cite{caesar2020nuscenes}.} 
    \name{}+ denotes the adoption of the ego planner input.
    ${}\ddagger$: The model is trained using only the trajectory prediction task for open-loop planning, without utilizing our generated OmniDrive Q\&A data.
    Methods marked as \textcolor{gray}{Hybrid Paradigm} here stack traditional and VLM-based approaches, and are thus incomparable.
    Results are highlighted in \textbf{bold} and \underline{underline} for the best and the second-best performance among VLM-based methods.
    }
    \label{tab:supp_open_loop_benchmark}
    \centering
    \resizebox{\linewidth}{!}{
    \begin{tabular}{l|cc|cccc|cccc|cccc|cccc}
    \toprule
    \multirow{3}{*}{Method} &
    \multicolumn{2}{c|}{\multirow{2}{*}{Ego Status}} & 
    \multicolumn{8}{c|}{ST-P3 Metrics} & 
    \multicolumn{8}{c}{UniAD Metrics} \\
    & \multicolumn{2}{c|}{} &
    \multicolumn{4}{c|}{L2 (m) $\downarrow$} & 
    \multicolumn{4}{c|}{Collision (\%) $\downarrow$} &
    \multicolumn{4}{c|}{L2 (m) $\downarrow$} & 
    \multicolumn{4}{c}{Collision (\%) $\downarrow$} \\
    & BEV & Planner & 1s & 2s & 3s &\cellcolor{gray!30}Avg. & 1s & 2s & 3s& 
    \cellcolor{gray!30}Avg. & 1s & 2s & 3s &\cellcolor{gray!30}Avg. & 1s & 2s & 3s &\cellcolor{gray!30}Avg. \\
    
    \midrule
    \multicolumn{19}{l}{\textit{Traditional Modular Paradigm}} \\
    \midrule
    
    ST-P3~\cite{hu2022st} & - & - & 1.33 & 2.11 & 2.90 & \cellcolor{gray!30}2.11 & 0.23 & 0.62 & 1.27 & \cellcolor{gray!30}0.71 & 1.72 & 3.26 & 4.86 & \cellcolor{gray!30}3.28 & 0.44 & 1.08 & 3.01 & \cellcolor{gray!30}1.51 \\
    UniAD~\cite{hu2023planning}  & - & - & 0.44 & 0.67 & 0.96 & \cellcolor{gray!30}0.69 & 0.04 & 0.08 & 0.23 & \cellcolor{gray!30}0.12 & 0.48 & 0.96 & 1.65 & \cellcolor{gray!30}1.03 & 0.05 & 0.17 & 0.71 & \cellcolor{gray!30}0.31 \\
    VAD-Base~\cite{jiang2023vad} & - & - & 0.41 & 0.70 & 1.05 & \cellcolor{gray!30}0.72 & 0.07 & 0.17 & 0.41 & \cellcolor{gray!30}0.22 & 0.54 & 1.15 & 1.98 & \cellcolor{gray!30}1.22 & 0.10 & 0.24 & 0.96 & \cellcolor{gray!30}0.43 \\
    UAD~\cite{guo2025end} & - & - & 0.28 & 0.41 & 0.65 & \cellcolor{gray!30}0.45 & 0.01 & 0.03 & 0.14 & \cellcolor{gray!30}0.06 & 0.39 & 0.80 & 1.50 & \cellcolor{gray!30}0.90 & 0.01 & 0.12 & 0.43 & \cellcolor{gray!30}0.19 \\
    MomAD~\cite{song2025don} & - & - & 0.28 & 0.49 & 0.78 & \cellcolor{gray!30}0.52 & 0.08 & 0.14 & 0.34 & \cellcolor{gray!30}0.19 & 0.36 & 0.83 & 1.56 & \cellcolor{gray!30}0.91 & 0.06 & 0.23 & 1.00 & \cellcolor{gray!30}0.43 \\
    GenAD~\cite{zheng2024genad} & - &\cmark & 0.31 & 0.57 & 0.91 & \cellcolor{gray!30}0.60 & 0.01 & 0.05 & 0.22 & \cellcolor{gray!30}0.09 & 0.43 & 0.88 & 1.62 & \cellcolor{gray!30}0.98 & 0.06 & 0.16 & 0.68 & \cellcolor{gray!30}0.30 \\
    Drive-WM~\cite{wang2024driving} &\cmark &\cmark & 0.43 & 0.77 & 1.20 & \cellcolor{gray!30}0.80 & 0.10 & 0.21 & 0.48 & \cellcolor{gray!30}0.26 & - & - & - & \cellcolor{gray!30}- & - & - & - & \cellcolor{gray!30}- \\
    SparseDrive~\cite{sun2025sparsedrive} & - &\cmark & 0.29 & 0.55 & 0.91 & \cellcolor{gray!30}0.58 & 0.01 & 0.02 & 0.13 & \cellcolor{gray!30}0.06 & 0.44 & 0.92 & 1.69 & \cellcolor{gray!30}1.01 & 0.07 & 0.19 & 0.71 & \cellcolor{gray!30}0.32 \\
    DiffusionDrive~\cite{liao2025diffusiondrive} & - &\cmark & 0.27 & 0.54 & 0.90 & \cellcolor{gray!30}0.57 & 0.03 & 0.05 & 0.16 & \cellcolor{gray!30}0.08 & - & - & - & \cellcolor{gray!30}- & - & - & - & \cellcolor{gray!30}- \\    
    
    \midrule
    \multicolumn{19}{l}{\textit{VLM-based Paradigm}} \\
    \midrule
    
    EMMA~\cite{hwang2024emma} & - & - & \textbf{0.14} & \textbf{0.29} & \underline{0.54} & \cellcolor{gray!30}\textbf{0.32} & - & - & - & \cellcolor{gray!30}- & - & - & - & \cellcolor{gray!30}- & - & - & - & \cellcolor{gray!30}- \\
    RDA-Driver~\cite{huang2024making} &\cmark &\cmark & 0.17 & 0.37 & 0.69 & \cellcolor{gray!30}0.40 & \textbf{0.01} & \textbf{0.05} & \underline{0.26} & \cellcolor{gray!30}\textbf{0.10} & \underline{0.23} & \underline{0.73} & \underline{1.54} & \cellcolor{gray!30}\underline{0.80} & \textbf{0.00} & \textbf{0.13} & \underline{0.83} & \cellcolor{gray!30}\textbf{0.32} \\
    DriveVLM~\cite{tian2025drivevlm} & - & \cmark & 0.18 & 0.34 & 0.68 & \cellcolor{gray!30}0.40 & - & - & - & \cellcolor{gray!30}- & - & - & - & \cellcolor{gray!30}- & - & - & - & \cellcolor{gray!30}- \\
    ORION~\cite{fu2025orion}  &\cmark & - & 0.17 & \underline{0.31} & 0.55 & \cellcolor{gray!30}0.34  & - & - & - & \cellcolor{gray!30}- & - & - & - & \cellcolor{gray!30}- & - & - & - & \cellcolor{gray!30}- \\
    OmniDrive-Q~\cite{wang2025omnidrive} & - & - & 1.15 & 1.96 & 2.84 & \cellcolor{gray!30}1.98 & - & - & - & \cellcolor{gray!30}- & - & - & - & \cellcolor{gray!30}- & - & - & - & \cellcolor{gray!30}- \\
    OmniDrive-Q++~\cite{wang2025omnidrive} &\cmark &\cmark & \textbf{0.14} & \textbf{0.29} & 0.55 & \cellcolor{gray!30}\underline{0.33} & - & - & - & \cellcolor{gray!30}- & - & - & - & \cellcolor{gray!30}- & - & - & - & \cellcolor{gray!30}- \\
    OmniDrive-L$\ddagger$~\cite{wang2025omnidrive}  & -& -& 1.47 & 2.43 & 3.38 & \cellcolor{gray!30}2.43 & - & - & - & \cellcolor{gray!30}- & - & - & - & \cellcolor{gray!30}- & - & - & - & \cellcolor{gray!30}- \\
    OmniDrive-L++$\ddagger$~\cite{wang2025omnidrive}  & - &\cmark & 0.31 & 0.62 & 1.06 & \cellcolor{gray!30}0.66 & - & - & - & \cellcolor{gray!30}- & - & - & - & \cellcolor{gray!30}- & - & - & - & \cellcolor{gray!30}- \\
    OmniDrive-L~\cite{wang2025omnidrive}  & - & - & 1.43 & 2.34 & 3.24 & \cellcolor{gray!30}2.34 & - & - & - & \cellcolor{gray!30}- & - & - & - & \cellcolor{gray!30}- & - & - & - & \cellcolor{gray!30}- \\
    OmniDrive-L++~\cite{wang2025omnidrive} & - &\cmark & \underline{0.15} & 0.36 & 0.70 & \cellcolor{gray!30}0.40 & - & - & - & \cellcolor{gray!30}- & - & - & - & \cellcolor{gray!30}- & - & - & - & \cellcolor{gray!30}- \\

    \midrule


    \name{}~(\textbf{ours}) & - & - & 1.06 & 1.79 & 2.55 & \cellcolor{gray!30}1.80 & 0.35 & 0.61  & 1.31 & \cellcolor{gray!30}\underline{0.76} & 1.41 & 2.88 & 4.51 & \cellcolor{gray!30}2.93 & 0.59 & 1.72 & 4.53 & \cellcolor{gray!30}2.28 \\
    \name{}+~(\textbf{ours}) & - & \cmark & \underline{0.15} & \textbf{0.29} & \textbf{0.51} & \cellcolor{gray!30}\textbf{0.32} & \underline{0.05} & \underline{0.08} & \textbf{0.16} & \cellcolor{gray!30}\textbf{0.10} & \textbf{0.20} & \textbf{0.53} & \textbf{1.13} & \cellcolor{gray!30}\textbf{0.62} & \underline{0.10} & \underline{0.31} & \textbf{0.80} & \cellcolor{gray!30}\underline{0.40} \\

    \midrule
    \multicolumn{19}{l}{\textcolor{gray}{\textit{Hybrid Paradigm}}} \\
    \midrule

    \textcolor{gray}{VLP~\cite{pan2024vlp}} & \textcolor{gray}{\cmark} & \textcolor{gray}{-} & \textcolor{gray}{0.30} & \textcolor{gray}{0.53} & \textcolor{gray}{0.84} & \cellcolor{gray!30}\textcolor{gray}{0.55} & \textcolor{gray}{0.01} & \textcolor{gray}{0.07} & \textcolor{gray}{0.38} & \cellcolor{gray!30}\textcolor{gray}{0.15} & \textcolor{gray}{0.36} & \textcolor{gray}{0.68} & \textcolor{gray}{1.19} & \cellcolor{gray!30}\textcolor{gray}{0.74} & \textcolor{gray}{0.03} & \textcolor{gray}{0.12} & \textcolor{gray}{0.32} & \cellcolor{gray!30}\textcolor{gray}{0.16} \\
    \textcolor{gray}{ReAL-AD~\cite{lu2025real}} & \textcolor{gray}{\cmark} & \textcolor{gray}{-} & \textcolor{gray}{0.30} & \textcolor{gray}{0.48} & \textcolor{gray}{0.67} & \cellcolor{gray!30}\textcolor{gray}{0.48} & \textcolor{gray}{0.07} & \textcolor{gray}{0.10} & \textcolor{gray}{0.28} & \cellcolor{gray!30}\textcolor{gray}{0.15} & \textcolor{gray}{0.40} & \textcolor{gray}{0.71} & \textcolor{gray}{1.14} & \cellcolor{gray!30}\textcolor{gray}{0.77} & \textcolor{gray}{0.02} & \textcolor{gray}{0.12} & \textcolor{gray}{0.37} & \cellcolor{gray!30}\textcolor{gray}{0.17} \\
    \textcolor{gray}{DriveVLM-Dual~\cite{tian2025drivevlm}} & \textcolor{gray}{\cmark} & \textcolor{gray}{-} & \textcolor{gray}{0.15} & \textcolor{gray}{0.29} & \textcolor{gray}{0.48} & \cellcolor{gray!30}\textcolor{gray}{0.31} & \textcolor{gray}{0.05} & \textcolor{gray}{0.08} & \textcolor{gray}{0.17} & \cellcolor{gray!30}\textcolor{gray}{0.10} & \textcolor{gray}{-} & \textcolor{gray}{-} & \textcolor{gray}{-} & \cellcolor{gray!30}\textcolor{gray}{-} & \textcolor{gray}{-} & \textcolor{gray}{-} & \textcolor{gray}{-} & \cellcolor{gray!30}\textcolor{gray}{-} \\
    \textcolor{gray}{SOLVE-VLM~\cite{chen2025solve}} & \textcolor{gray}{\cmark} & \textcolor{gray}{-} & \textcolor{gray}{0.13} & \textcolor{gray}{0.25} & \textcolor{gray}{0.47} & \cellcolor{gray!30}\textcolor{gray}{0.28} & \textcolor{gray}{-} & \textcolor{gray}{-} & \textcolor{gray}{-} & \cellcolor{gray!30}\textcolor{gray}{-} & \textcolor{gray}{-} & \textcolor{gray}{-} & \textcolor{gray}{-} & \cellcolor{gray!30}\textcolor{gray}{-} & \textcolor{gray}{-} & \textcolor{gray}{-} & \textcolor{gray}{-} & \cellcolor{gray!30}\textcolor{gray}{-} \\
    \textcolor{gray}{Senna~\cite{jiang2024senna}} & \textcolor{gray}{\cmark} & \textcolor{gray}{-} & \textcolor{gray}{0.11} & \textcolor{gray}{0.21} & \textcolor{gray}{0.35} & \cellcolor{gray!30}\textcolor{gray}{0.22} & \textcolor{gray}{0.04} & \textcolor{gray}{0.08} & \textcolor{gray}{0.13} & \cellcolor{gray!30}\textcolor{gray}{0.08} & \textcolor{gray}{-} & \textcolor{gray}{-} & \textcolor{gray}{-} & \cellcolor{gray!30}\textcolor{gray}{-} & \textcolor{gray}{-} & \textcolor{gray}{-} & \textcolor{gray}{-} & \cellcolor{gray!30}\textcolor{gray}{-} \\
    \bottomrule
    \end{tabular}}
\end{table*}

\begin{table}[t]
    \caption{
    \textbf{Closed-loop planning results on Bench2Drive~\cite{jia2024bench2drive}.}
    Results are highlighted in \textbf{bold} and \underline{underline} for the best and the second-best performance among VLM-based methods.
    }
    \label{tab:supp_closed_loop_benchmark}
    \centering
    \resizebox{\linewidth}{!}{
    \begin{tabular}{l|cc}
    \toprule
    \multirow{2}{*}{Method} &
    \multicolumn{2}{c}{Closed-loop Metric} \\
    & Driving Score $\uparrow$ & Success Rate(\%) $\uparrow$ \\
    
    \midrule
    \multicolumn{3}{l}{\textit{Traditional Modular Paradigm}} \\
    \midrule
    
    AD-MLP~\cite{zhai2023rethinking} & 18.05 & 0.00 \\ 
    UniAD-Base~\cite{hu2023planning} & 45.81 & 16.36 \\ 
    VAD-Base~\cite{jiang2023vad} & 42.35 & 15.00 \\ 
    MomAD~\cite{song2025don} & 44.54 & 16.71 \\ 
    GenAD~\cite{zheng2024genad} & 44.81 & 15.90 \\ 
    SparseDrive~\cite{sun2025sparsedrive} & 47.38 & 17.72 \\ 
    UAD~\cite{guo2025end} & 49.22 & 20.45 \\ 
    SeerDrive~\cite{zhangfuture} & 58.32 & 30.17 \\ 
    WoTE~\cite{li2025end} & 61.71 & 31.36 \\ 
    DriveDPO~\cite{shang2025drivedpo} & 62.02 & 30.62 \\ 
    ThinkTwice~\cite{jia2023think} & 62.44 & 37.17 \\ 
    DriveTransformer-L~\cite{jia2025drivetransformer} & 63.46 & 38.60 \\ 
    DriveAdapter~\cite{jia2023driveadapter} & 64.22 & 42.08 \\ 
    GEMINUS~\cite{wan2025geminus} & 65.39 & 37.73 \\ 
    Raw2Drive~\cite{yang2025raw2drive} & 71.36 & 50.24 \\ 
    Hydra-NeXt~\cite{li2025hydra} & 73.86 & 53.22 \\ 
    DiffusionDrive~\cite{liao2025diffusiondrive} & 77.68 & 52.72 \\ 
    PGS~\cite{huangprioritizing} & 78.08 & 48.64 \\ 
    GaussianFusion~\cite{liu2025gaussianfusion} & 79.10 & 54.40 \\ 
    TF++~\cite{zimmerlin2024hidden} & 84.21 & 64.39 \\ 
    R2SE~\cite{liu2025reinforced} & 86.28 & 67.76 \\ 
    HiP-AD~\cite{tang2025hipad} & 86.77 & 69.09 \\ 
    AlignDrive~\cite{wu2026aligndrive} & 89.07 & 73.18 \\
     
    \midrule
    \multicolumn{3}{l}{\textit{VLM-based Paradigm}} \\
    \midrule
    
    ReAL-AD~\cite{lu2025real} & 41.17 & 11.36 \\ 
    Dual-AEB~\cite{zhang2025dual} & 45.23 & 10.00 \\ 
    X-Driver~\cite{liu2025x} & 51.70 & 18.10 \\ 
    VDRive~\cite{guo2025vdrive} & 66.25 & 50.51 \\ 
    StuckSolver~\cite{bao2025large} & 70.89 & 50.01 \\ 
    DriveMoE~\cite{yang2025drivemoe} & 74.22 & 48.64 \\ 
    ETA~\cite{hamdan2025eta} & 74.33 & 48.33 \\ 
    VLR-Drive~\cite{kong2025vlr} & 75.01 & 50.00 \\ 
    ORION~\cite{fu2025orion} & 77.74 & 54.62 \\ 
    SimLingo~\cite{renz2025simlingo} & \textbf{85.07} & \textbf{67.27} \\ 

    \midrule

    \name{}+~(\textbf{ours}) & \underline{78.02} & \underline{55.11} \\ 
    
    \bottomrule
    \end{tabular}}
    \vspace{-5pt}
\end{table}

Constrained by the limited space in the main paper, we list only the primary relevant works in the benchmark comparisons. 
Therefore, we provide more comprehensive benchmark comparisons for open-loop and closed-loop planning in~\cref{tab:supp_open_loop_benchmark} and~\cref{tab:supp_closed_loop_benchmark}, respectively. 
It is worth noting that existing nuScenes~\cite{caesar2020nuscenes} open-loop evaluations utilize differing sets of metrics in different studies. 
While the main paper employs the OmniDrive~\cite{wang2025omnidrive} version commonly used by VLM-based frameworks, \Cref{tab:supp_open_loop_benchmark} provides results derived using the evaluation metrics from ST-P3~\cite{hu2022st} and UniAD~\cite{hu2023planning}.

\section{Additional Visualization}
\label{sec:addl_vis}

\subsection{More Adaptability Analysis}
\label{subsec:addl_adaptability}

\begin{figure}[t]
    \centering
    \includegraphics[width=\linewidth]{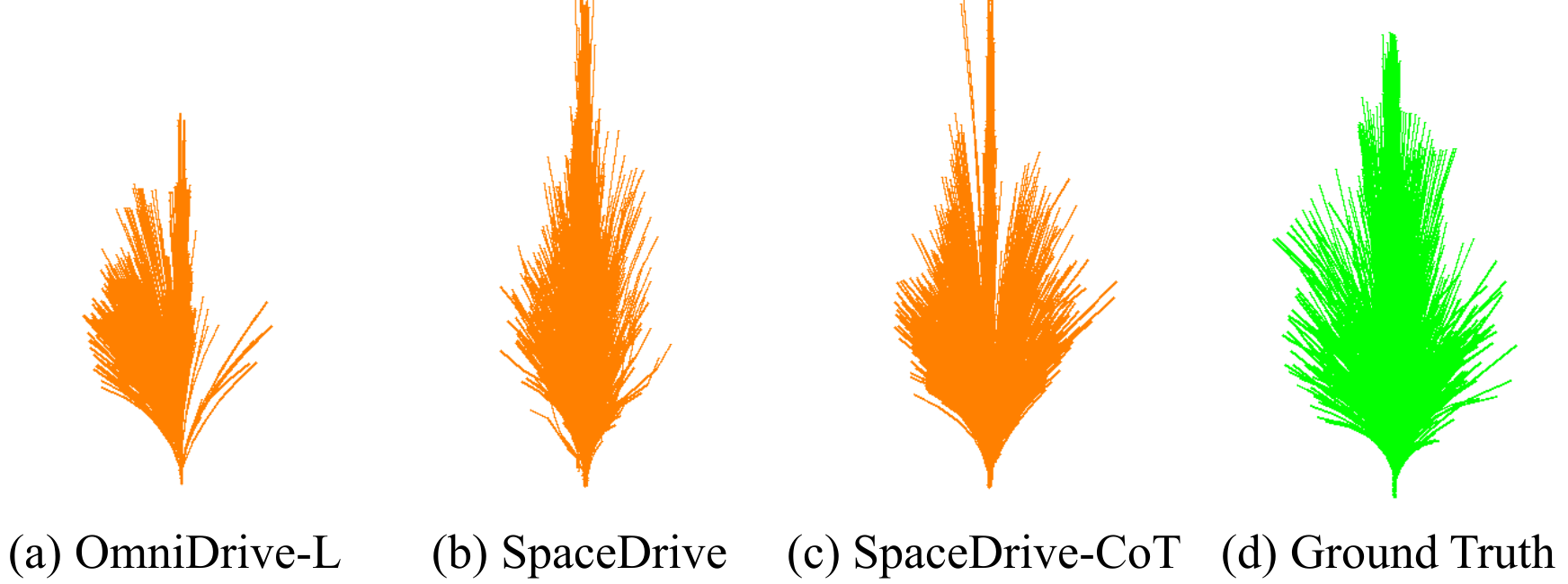}
    \caption{
    \textbf{Trajectory distribution of open-loop planning for different frameworks and ground truth.}
    }
    \label{fig:supp_mode_collapse}
\end{figure}

\Cref{fig:supp_mode_collapse} illustrates the distribution of planned trajectories across all scenarios under the open-loop setting of nuScenes~\cite{caesar2020nuscenes}.
We first analyze the output trajectory distribution of OmniDrive-L~\cite{wang2025omnidrive}, a typical scheme utilizing textual digit tokens for waypoint coordinates, shown in~\cref{fig:supp_mode_collapse}.a. 
Due to the VLM's limitations in numerical processing, as discussed in~\Cref{sec:intro}, OmniDrive-L exhibits clear mode collapse for right-turn cases. 
In sharp contrast, our \name{}, which is based on the universal 3D PE representation, significantly mitigates this issue, as shown in~\cref{fig:supp_mode_collapse}.b. 
Furthermore, when adopting inference techniques such as Chain-of-Thought during inference, the output trajectory planning demonstrates enhanced robustness (\cref{fig:supp_mode_collapse}.c) and closer alignment with the ground truth distribution (\cref{fig:supp_mode_collapse}.d). 
This result further supports the strong adaptability of our method to language model inference techniques.

\subsection{Failure Analysis of Textual Coordinate Output}
\label{subsec:omnidrive_failure_ana}

\begin{figure*}[t]
    \centering
    \includegraphics[width=\linewidth]{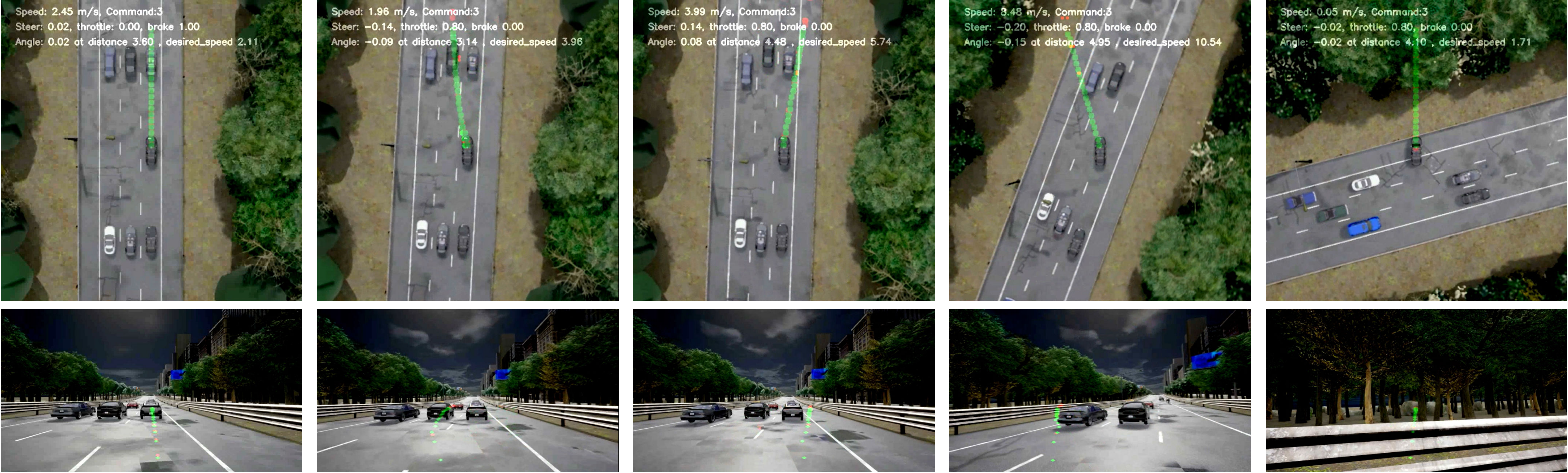}
    \caption{
    \textbf{Qualitative results of OmniDrive-L~\cite{wang2025omnidrive} in closed-loop setting on Bench2Drive~\cite{jia2024bench2drive}.} 
    \textcolor{green}{Green} and \textcolor{pink}{pink} dots represent path and speed waypoints, respectively. 
    Parameters such as speed and steering wheel angle can be found in the figures.
    }
    \label{fig:omnil_planning_vis}
    \vspace{-5pt}
\end{figure*}

A further quantitative analysis is conducted to assess the driving capability of conventional VLM-based models that output trajectory coordinates as textual digit tokens in closed-loop simulation, as shown in~\cref{fig:omnil_planning_vis}. 
We the exact same scenario as in~\cref{fig:planning_vis} and employ OmniDrive-L~\cite{wang2025omnidrive}, a framework structurally analogous to \name{}, utilizing the same closed-loop training configuration as in~\cref{subsec:training_details}. 
This figure clearly illustrates that in the closed-loop setting, the planned trajectories generated by OmniDrive-L collapse into an approximately straight line, and the directional control exhibits random oscillation. 
This phenomenon aligns with the mode collapse previously observed during open-loop evaluation (See \cref{subsec:addl_adaptability}). 
Critically, this oscillation is amplified over time, leading to vehicle instability and ultimately making the vehicle veer off the road and collide with the guardrail. 
This result provides strong empirical support for our analysis in~\cref{subsec:quantitative_results}: purely text-based trajectory coordinate output from VLMs is inadequate for reliable closed-loop driving.

\subsection{More Qualitative Closed-Loop Results}
\label{subsec:addl_qualitative_results}

\begin{figure*}[t]
    \centering
    \includegraphics[width=\linewidth]{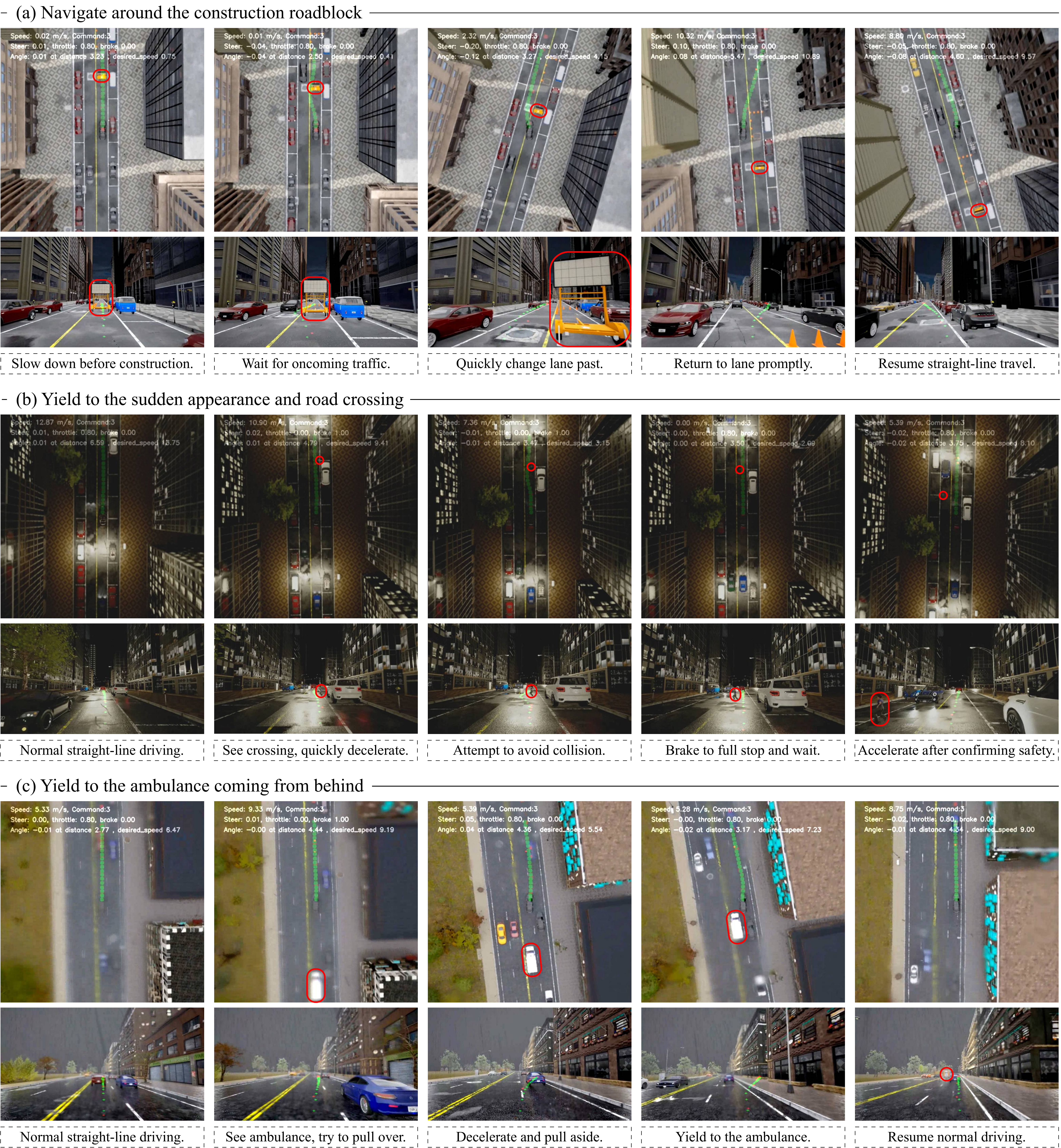}
    \caption{
    \textbf{More qualitative results of closed-loop evaluation on Bench2Drive~\cite{jia2024bench2drive}.} 
    We include 3 scenarios here to demonstrate the closed-loop planning capability of \name{}: (a) urban road construction; (b) a sudden pedestrian crossing; (c) yielding to an ambulance.
    \textcolor{green}{Green} and \textcolor{pink}{pink} dots represent path and speed waypoints, respectively. 
    \textcolor{red}{Red circles} indicate objects requiring attention in the scenario.
    Parameters such as speed and steering wheel angle can be found in the figures.
    }
    \label{fig:supp_planning_vis}
\end{figure*}

We present additional closed-loop simulation visualizations for \name{} in~\cref{fig:supp_planning_vis}, covering 3 representative safety-critical scenarios: (a) navigating around a construction zone requiring a brief excursion into the oncoming lane; (b) decelerating and yielding due to a sudden pedestrian crossing during normal driving; and (c) performing an emergency stop and yielding to an ambulance rapidly approaching from the rear.
All these scenarios demand the model to quickly establish a deep understanding of the 3D spatial context and generate a sound trajectory in a minimal timeframe. 
The visualizations clearly indicate that our proposed framework, by leveraging its unified 3D representation, effectively manages these critical, unforeseen situations. 
This further substantiates the efficacy of our proposed \name{} framework.

\end{document}